\documentclass[10pt,twocolumn,letterpaper]{article}

\usepackage{cvpr}              

\usepackage{multirow}
\usepackage{float} 
\usepackage{xcolor}
\usepackage{appendix}

\definecolor{gold}{RGB}{212,175,55}
\definecolor{silver}{RGB}{192,192,192}
\definecolor{bronze}{RGB}{205,127,50}

\definecolor{cvprblue}{rgb}{0.21,0.49,0.74}
\usepackage[pagebackref,breaklinks,colorlinks,allcolors=cvprblue]{hyperref}

\def\confName{CVPR}
\def\confYear{2026}

\title{FlexAvatar: Flexible Large Reconstruction Model for Animatable 
\\
Gaussian Head Avatars with Detailed Deformation
}

\author{Cheng Peng$^{1, 2, *}$, Zhuo Su$^{2, *, \dagger}$, Liao Wang$^{2, *}$, Chen Guo$^{1,2}$, Zhaohu Li$^{2}$, Chengjiang Long$^{2}$, \\  Zheng Lv$^{2}$, Jingxiang Sun$^{1}$, Chenyangguang Zhang$^{1}$, Yebin Liu$^{1, \dagger}$ \\ \\
$^{1}$Tsinghua University~~  $^{2}$ByteDance}

\begin{document}


\twocolumn[{
\maketitle
\begin{center}
    \captionsetup{type=figure}
    \vspace{-1mm}
    \includegraphics[width=1.\textwidth]{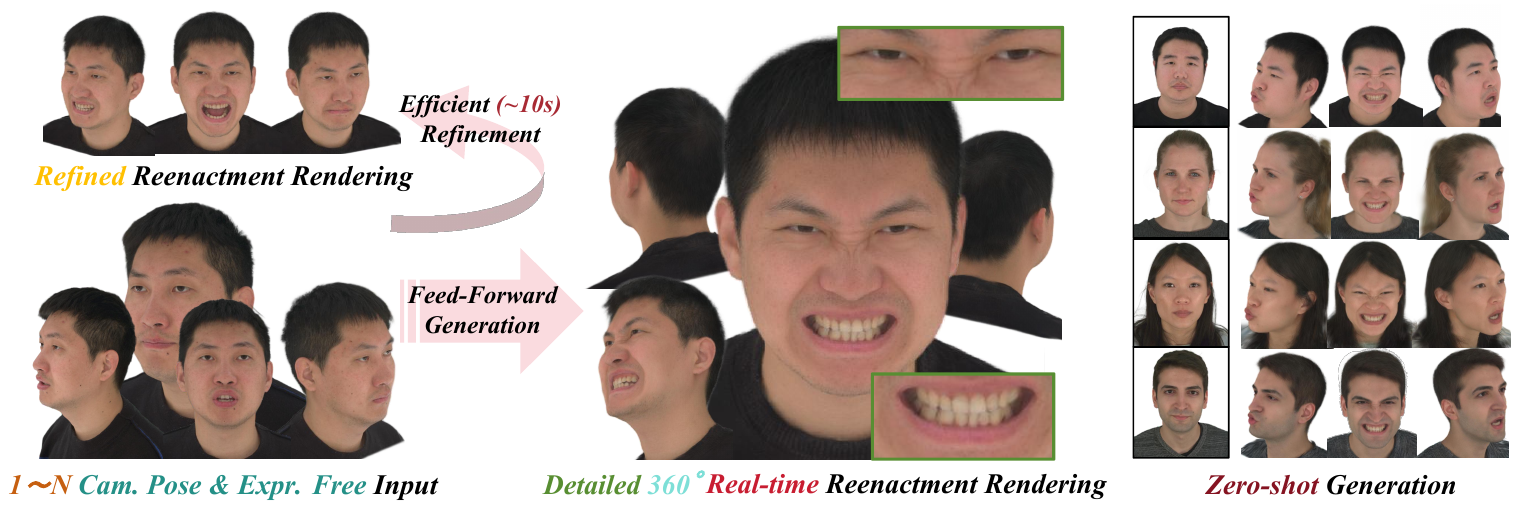}
    \vspace{-1mm}
    \captionof{figure}{
    Given single or sparse input images ({\em e.g.}, 1-4 images) \textbf{without camera poses or expression labels}, our proposed FlexAvatar produces \textbf{detailed, real-time 360°} reenactment renderings. Our feed-forward model already delivers high fidelity zero-shot results across diverse identities. An efficient refinement strategy (10 seconds) can be adopted to better match the input images.}
\end{center}
}]

\let\thefootnote\relax\footnotetext{$^*$ Equal contribution, $^\dagger$ Corresponding author.}


\begin{abstract}

\vspace{-4mm}

%
We present FlexAvatar, a flexible large reconstruction model for high-fidelity 3D head avatars with detailed dynamic deformation from single or sparse images, without requiring camera poses or expression labels. 
%
It leverages a transformer-based reconstruction model with structured head query tokens as canonical anchor to aggregate flexible input-number-agnostic, camera-pose-free and expression-free inputs into a robust canonical 3D representation.
For detailed dynamic deformation, we introduce a lightweight UNet decoder conditioned on UV-space position maps, which can produce detailed expression-dependent deformations in real time. 
To better capture rare but critical expressions like wrinkles and bared teeth, we also adopt a data distribution adjustment strategy during training to balance the distribution of these expressions in the training set.
Moreover, a lightweight 10-second refinement can further enhances identity-specific details in extreme identities without affecting deformation quality.
Extensive experiments demonstrate that our FlexAvatar achieves superior 3D consistency, detailed dynamic realism compared with previous methods, providing a practical solution for animatable 3D avatar creation. Our project page:  \href{https://pengc02.github.io/flexavatar}{pengc02.github.io/flexavatar} 

\end{abstract}    
\vspace{-15pt}
\section{Introduction}
\label{sec:intro}

The creation of photorealistic, animatable 3D head avatars is a long-standing goal for immersive communication, telepresence, and digital human applications.
With the emergence of 3D Gaussian Splatting (3DGS), real-time rendering of complex 3D scenes has become practical, spurring a new wave of avatar reconstruction methods~\cite{saito2024relightable, qian2024gaussianavatars,xiang2024flashavatar,guo2025segadrivable3dgaussian,li2025rgbavatar,Chen_2025_CVPR,HRAvatar,zhang2025fate}. 

Despite rapid progress, most existing pipelines remain constrained by multi-view captures, camera pose or expression inputs, which hinder scalability to diverse identities and casual users.
Recent works have explored different directions to mitigate these limitations.
2D-driven generators (e.g., GAGAvatar~\cite{chu2024gagavatar}, Portrait4D~\cite{deng2024portrait4d}) can synthesize visually appealing portraits, but they struggle to maintain 3D geometry consistency and fail to faithfully reproduce detailed dynamic deformations.
3D prior–based systems (e.g., HeadGAP~\cite{zheng2024headgap}, One2Avatar~\cite{yu2024one2avatar}) enforce geometric coherence but are limited by the small diversity of available 3D data and often rely on time-consuming inversion or subject-specific fine-tuning, which hinders scalability and fast deployment.
Large reconstruction or foundation models (e.g., LRM\cite{hong2023lrm}, LAM\cite{he2025lam}, Avat3R\cite{Kirschstein2025avat3r}) achieve strong generalization via data and model scaling. However, they are typically limited to single images or a fixed number of inputs, and their outputs often fail to fully match driving signals or capture fine-grained dynamic deformations. Moreover, many rely on cross-attention for expression changes, hindering real-time deployment.

In this paper, to overcome input constraints, weak expression fidelity and pseudo-3D artifacts, we present FlexAvatar, a flexible large reconstruction model for detailed Gaussian head avatars, generalizing from single or sparse inputs without camera poses or expression labels. This differs from existing pipelines which are either limited to single-image reconstruction or rely on multi-view inputs with known camera parameters and consistent expressions, making them neither pose-free nor expression-free.

\begin{figure*}[h]
\vspace{-2pt}
    \centering
    \includegraphics[width=\linewidth]{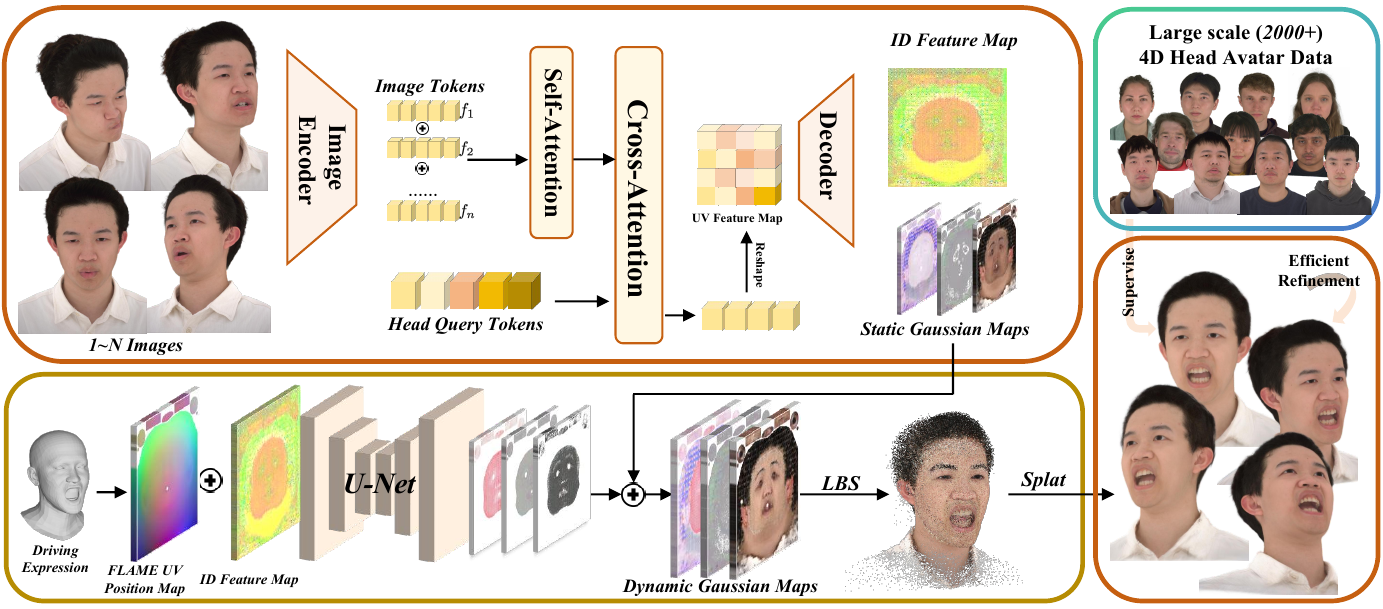}
    \vspace{-2mm}
    \caption{FlexAvatar reconstructs a high-quality Gaussian head avatar by mapping a flexible number of input images with varying expressions and camera views into Gaussian representations in UV space. We use a flexible feed-forward backbone to obtain static Gaussian attributes and an identity feature map from input
    images. Given a driving expression signal, we then convert it into a FLAME UV position map and concatenate it with the backbone's identity feature map; to support real-time driving and produce high-quality dynamic results, the concatenated representation is then fed into a UNet to generate expression-dependent dynamic Gaussian attributes in UV space, which are then sampled into FLAME space with LBS for rendering. Optionally, an efficient refinement can further improve the results.
    }
    \label{fig:overview}
    \vspace{-4mm}
\end{figure*}

As illustrated in Fig.~\ref{fig:overview}, FlexAvatar's feed-forward architecture comprises a transformer-based reconstruction model aggregating arbitrary views into a canonical head representation via structured Head Query tokens, and a lightweight UNet-based Dynamic Decoder producing expression-dependent Gaussian deformations in real time to capture fine-grained details. The framework is trained in an end-to-end manner with a data distribution adjustment strategy emphasizing challenging expressions (e.g., wrinkles, bared teeth). This formulation surpasses existing LRM-based methods in geometry consistency, identity fidelity, and dynamic expression realism.




While our feed-forward Gaussian avatar generation is inherently scalable and can improve by absorbing more data, it struggles with long-tail traits and challenging appearances due to limited high-quality 3D head data. To remedy this, we introduce a lightweight test-time refinement that uses only the input images. Taking roughly 10 seconds, it selectively enhances identity-specific details in these difficult regions, complementing the scalable backbone in data-scarce settings.
This design effectively combines the scalability and robustness of large reconstruction models with the efficiency of Gaussian representations, providing a practical foundation for real-time, animatable 3D avatars.

To summarize, our main contributions are as follows:

\begin{itemize}
\item \textbf{Flexible Reconstruction Model.} 
We propose the first camera-pose-, expression-, and input-count-free Gaussian head avatar framework, where Structured Head Query tokens with transformer aggregation enable robust 3D reconstruction from 1 to N input images.

\item \textbf{Dynamic Gaussian Deformation decoding.} 
We introduce a lightweight UNet-based dynamic decoder that leverages UV-space position maps to learn spatially aligned, expression-dependent Gaussian deformations, enabling real-time and geometrically consistent dynamic details.

\item \textbf{Distribution Adjusted dynamic learning.} 
We rebalance the data by selecting expressive anchors and retrieving similar frames via FLAME cosine similarity, plus random samples per ID, enabling faster convergence and more realistic dynamics.

\item \textbf{Superior Performance to SOTA Methods.} 
Our proposed feed-forward model achieves high-fidelity reconstruction and dynamic realism, surpassing prior LRM-based avatars. An efficient refinement further boosts appearance and identity without impacting runtime. Experiments show our FlexAvatar sets a new state of the art.

\end{itemize}
\section{Related Work}
\label{sec:related_works}

\subsection{3D-aware Portrait Animation}


Recent research on 3D-aware portrait animation has evolved along two primary technical tracks. The first track focuses on 2D generation and driven under 3D control signals, exemplified by methods that leverage personalized diffusion models~\cite{taubner2024cap4d,taubner2025mvp4d,zhao2025x}. While effective for frontal views, these approaches often rely on 2D screen-space neural renderers, prioritizing visual fidelity over strict 3D consistency. The second track employs explicit 3D representations for face animation, where the face is modeled using 3D volumes~\cite{drobyshev2022megaportraits}, neural pipelines~\cite{tewari2020stylerig}, meshes~\cite{khakhulin2022rome}, TriPlane structures~\cite{deng2024portrait4d, deng2024portrait4dv2, chu2024gpavatar, tran2024voodoo3d, tran2024voodooxp}, or 3D Gaussian primitives~\cite{chu2024gagavatar,he2025lam}. In most of these systems, motion and expression are typically driven by a 3D morphable model (3DMM)~\cite{blanz19993dmm, li2017flame}.
Although these approaches can animate an individual portrait, they are primarily tailored to synthesizing images from a frontal viewpoint and often employ 2D screen-space neural renderers, which tend to prioritize visual fidelity over strict 3D consistency.

\subsection{Photo-realistic 3D Avatar}

Existing methods can learn high-quality drivable avatars from multi-view \cite{qian2024gaussianavatars,xu2023gaussianheadavatar,zielonka2023insta,zielonka2023d3ga,jiang2025reperformer, teotia2024gaussianheads} or monocular videos \cite{xiang2024flashavatar,zheng2023pointavatar,chen2024monogaussianavatar,zheng2022imavatar,xu2023avatarmav, zhang2025cad, gafni2021nerface,grassal2022nha,tang2024gaf,li2025rgbavatar, HRAvatar, peng2025parametricgaussianhumanmodel}. However, these per-subject approaches require lengthy training and over-rely on interpolating within their training data, hindering generalization to unseen expressions or identities. The availability of large-scale multi-view datasets \cite{kirschstein2023nersemble, ava256, pan2024renderme360, yang2020facescape, buehler2024cafca} has enabled models that capture strong facial priors. This includes NeRF-based parametric models \cite{hong2022headnerf,zheng2024ohta, yang2024vrmm,yu2024one2avatar}, 3D Gaussian-based methods \cite{xu2025gphm,zheng2024headgap, guo2025segadrivable3dgaussian, li2024uravatar}, and 3D GANs learned from 2D image collections \cite{sun2023next3d,tang20233dfaceshop,lee2022expgan,bergman2022gnarf,yu2025gaia,zhao2024invertavatar,chan2021pigan,chan2022eg3d,gu2021stylenerf}. Despite high visual quality, obtaining a personalized avatar often necessitates costly optimization \cite{sun2022ide3d,liu20223dfmgan,yuan2023goae,bhattarai2024triplanenet}. A fundamental limitation persists: even with more datasets, the number of distinct identities (often in the low thousands) is insufficient to capture real-world facial diversity. While 3D GANs alleviate data constraints, they can be brittle when extrapolating to novel viewpoints.


\subsection{Feed-forward 3D Reconstruction Models}
Recently, generalized feed-forward methods for sparse-view 3D reconstruction have attracted growing attention due to their effectiveness in under-constrained scenarios. For instance, the Large Reconstruction Model (LRM)~\cite{hong2023lrm,zhuang2024idolinstantphotorealistic3d} employs a transformer-based encoder–decoder architecture to reconstruct NeRF representations using only a single input image. Subsequent works have progressively transitioned toward generating 3D Gaussian representations conditioned on single or few-view inputs~\cite{tang2025lgm, xu2024grm, zhang2025gslrm, charatan2024pixelsplat, chen2025mvsplat, liu2025mvsgaussian, sun2025dreamcraft3d++, zou2024triplane, pan2024humansplat, zhao2025hunyuan3d, ye2025hi3dgen}, demonstrating impressive novel-view synthesis capabilities. Most recently, FaceLift~\cite{lyu2024facelift} integrates personalized multi-view diffusion with a large-scale reconstruction backbone, enabling detailed 3D head reconstruction from just one image. Additionally, the transformer-based paradigm has successfully extended from static inputs to monocular video sequences, achieving compelling 4D dynamic reconstructions~\cite{ren2024l4gm}.
\vspace{-6pt}
\section{Method}

FlexAvatar constructs detailed Gaussian head avatars from single or sparse images under unconstrained conditions, without requiring camera poses or expression annotations.
The framework follows a feed-forward large reconstruction model paradigm, {\em i.e.}, an attention-based transformer encoder generalizes across camera pose, expression, and identity variations, while a UNet-based decoder generates view-consistent, expression-aware appearance.

An optional test-time refinement step further improves personalization and handles challenging cases with highly personalized features, while maintaining real-time efficiency.
Fig.~\ref{fig:overview} provides an overview of our method.

\subsection{Flexible Reconstruction Model
}
The flexibility of FlexAvatar stems from the strong generalization ability of the attention mechanism, 
which unifies inputs of arbitrary numbers, camera poses, and expressions into a consistent canonical representation.
Each image is encoded by a frozen foundation vision transformer (DINOv3 \cite{simeoni2025dinov3}) $E(\cdot)$ to extract dense multi-scale features:
\begin{equation}
f_i = E(I_i), i \in \{1,\dots,N\},
\end{equation}
where the encoder $E(\cdot)$ extracts visual features $f_i \in \mathbb{R}^{L \times D}$ that implicitly capture camera view and expression variations, without requiring explicit camera or expression inputs.
During training, we randomly select 1 to 4 diverse viewpoints and four timesteps with varying expressions to form the input images.

\vspace{1mm}
\noindent\textbf{Input-Fusion for Number Flexibility.}
We use a number-agnostic fusion to project all images into a unified latent space as $\{f_i\}_{i=1}^N$  features and fused by a global self-attention layer:
\begin{equation}
F_{\text{agg}} = \text{SelfAttn}(f_1, f_2, \dots, f_N); F_{\text{agg}} \in \mathbb{R}^{(N \times L) \times D}.
\end{equation}
where the attention operation is naturally defined on variable-length token sequences of size $(N \times L)$.
It dynamically adjusts attention weights based on visual content, 
making the fusion process inherently input-count-agnostic and enabling the same network to fuse a single image or multiple views seamlessly.

\vspace{1mm}
\noindent\textbf{Head Query for Cam. Pose and Expr. Flexibility.
}
To achieve camera-pose- and expression-free modeling, we introduce a set of structured and learnable Head Query tokens ($Q_{\mathrm{H}}$) that serve as canonical anchors for aggregating the fused image features.
The tokens are optimized jointly with the network and aligned with the aggregated image features through several cross-attention layers, which transform variable-length fused inputs into a fixed-dimensional 3D head representation.
\begin{equation}
\begin{split}
F_{\mathrm{Q}} &= \operatorname{CrossAttn}\bigl(Q_{\mathrm{H}}, F_{\mathrm{agg}}\bigr) \\
&= \operatorname{softmax}\!\left(\frac{Q_{\mathrm{H}} K_{\mathrm{agg}}^{\top}}{\sqrt{D}}\right)V_{\mathrm{agg}}
\end{split}
\end{equation}
where $Q_{\mathrm{H}} \in \mathbb{R}^{N_H \times D}$ denotes the Head Query tokens,
and $K_{\mathrm{agg}}, V_{\mathrm{agg}} \in \mathbb{R}^{(N \times L) \times D}$ are the key and value projections of the aggregated image features $F_{\mathrm{agg}}$.
The resulting query token representation $F_{\mathrm{Q}} \in \mathbb{R}^{N_H \times D}$ encodes canonical, pose-free, and expression-free head features.
To impose canonical 3D constraints and link the implicit query token representations to spatial geometry, we reshape the aggregated query token features $F_{\mathrm{Q}}$ into a UV feature map:
\begin{equation}
F_{\mathrm{UV}} \in \mathbb{R}^{H \times W \times D}, N_H = H \cdot W,
\end{equation}
where $H$ and $W$ denote the height and width of the UV map. 
Correspondences are thus indexed in UV space and inferred purely from visual cues, without requiring camera poses or discrete expression labels, providing a unified representation that bridges input-count-agnostic 2D observations and 3D Gaussian decoding.

\vspace{1mm}
\noindent\textbf{Feature Decoding.} 
We then employ several convolutional heads as a decoder to transform the UV-space feature map into an ID feature map and static Gaussian attribute maps of the required spatial dimensions. Formally,
\begin{equation}
F_{\mathrm{id}}, G_{\mathrm{st}} = \operatorname{Decoder}\bigl(F_{\mathrm{UV}}\bigr),
\end{equation}
where the static Gaussian attribute maps $G_{\mathrm{st}} = \{P, \alpha, S, C, R\}$ denote position, opacity, color, scale and rotation maps in canonical UV space, which we use for rendering as standard 3D Gaussian splatting techniques~\cite{kerbl3Dgaussians}.
$F_{\mathrm{id}}$ contains discriminative identity features used for subsequent dynamic driving and synthesis.

\subsection{Dynamic Gaussian Deformation Decoding
}
\label{Decoding}

Driving detailed facial avatars with dynamic deformation remains challenging. Pure FLAME‑based driving, such as LAM~\cite{he2025lam}, often fails to reproduce fine-grained dynamic deformations, while lightweight MLP drivers~\cite{chu2024gagavatar} produce limited realism with noticeable artifacts, and heavy cross‑attention mechanisms~\cite{Kirschstein2025avat3r}, though expressive, are too costly for real-time use. To address this trade-off, we propose a driving paradigm that is both efficient and effective at preserving expressive dynamics, consisting of three key components: position map driving, UNet-based dynamic decoding and data distribution adjustment for training.

\vspace{1mm}
\noindent\textbf{Position Map as Driving Signals.} 
To capture subtle, localized dynamic deformations in facial avatars, we leverage UV position map of FLAME~\cite{li2017flame} template.
This map align the identity feature map above and directly encodes local vertex displacements, facilitating accurate and expressive dynamic deformation learning.
Specifically, given FLAME expression coefficients, we first construct a FLAME UV position map \(P_{\mathrm{driving}}\) by deforming template vertex coordinates according to the expression parameters and sampling the resulting 3D positions into UV space via barycentric interpolation. This map is concatenated with the identity feature map along the channel dimension:  
\begin{equation}
\tilde{F}_{UV} = F_{\mathrm{id}} \oplus P_{\mathrm{driving}}
\end{equation}
where \(\oplus\) denotes channel-wise concatenation.

\vspace{1mm}
\noindent\textbf{UNet-based Dynamic Decoding.} 
To accurately capture both global facial motion and fine-grained expression details guided by the UV position map, the driver must model local deformation while maintaining global consistency. 
We design a UNet-based driver that operates directly on 2D UV maps: it efficiently aggregates multi-scale local and global features, preserves spatial neighborhood information, and naturally aligns with the position map. Compared to lightweight MLP drivers, our approach better captures subtle dynamics such as wrinkles and eyelid motion while remaining computationally efficient for real-time animation.
Specifically, the dynamic delta Gaussian maps are decoded as: 

\begin{equation}
\Delta G_{\mathrm{dyn}} = \operatorname{UNet}(\tilde{F}_{UV}).
\end{equation}
These are then fused with the static Gaussian attributes \(G_{\mathrm{st}}\) within predefined dynamic regions:  
\begin{equation}
G_{\mathrm{dyn}} = G_{\mathrm{st}} + M_{\mathrm{dyn}} \odot \Delta G_{\mathrm{dyn}}
\end{equation}
where \(M_{\mathrm{dyn}}\in\{0,1\}^{H\times W}\) denotes the UV mask for dynamic regions (e.g., face, mouth, eyes), and \(\odot\) is the element-wise product.  

Finally, the updated dynamic Gaussian attributes are rendered using differentiable Gaussian splatting:  
\begin{equation}
I = \mathcal{R}(\mathrm{LBS}(G_{\mathrm{dyn}}), \Theta)
\end{equation}
where \(\mathrm{LBS}(\cdot)\) applies linear blend skinning of FLAME to Gaussian positions and rotations, \(\mathcal{R}(\cdot, \cdot)\) is the differentiable renderer, and \(\Theta\) denotes camera parameters.

\vspace{1mm}
\noindent\textbf{Data Distribution Adjustment.} 
Furthermore, during our training, we observed that standard training protocols are insufficient because datasets are dominated by neutral or transitional expressions, while rare but critical dynamics (e.g., wrinkles, bared teeth) appear infrequently. To address this imbalance, we introduce a distribution-adjustment scheme: we select 20 expressive anchor expressions and retrieve similar frames across all IDs using cosine similarity of FLAME coefficients, supplemented with 6 random expressions per ID, as shown in Fig.~\ref{fig:Anchor}.
As demonstrated in Fig.~\ref{fig:evaluation-fig}(c), the proposed distribution adjustment effectively mitigates the data imbalance by substantially increasing the prevalence of rare dynamic expressions. This strategic augmentation thereby provides the model with sufficient and diverse exemplars to learn a more comprehensive and robust dynamic prior.
Finally, each ID is then trained with 30 views over 26 timesteps, enabling more realistic dynamic rendering and faster-converging learning of these important expressions.

\vspace{1mm}
\subsection{Training Objectives}
As for training objectives, we supervise the predicted images with standard photometric and perceptual losses~\cite{zhang2018lpips}: 
\begin{equation}
\begin{aligned}
\mathcal{L}_{l1} & =\left\|I_{pred}-I_{g t}\right\|_1, \\
\mathcal{L}_{ssim} & =\operatorname{SSIM}\left(I_{pred}, I_{gt}\right),\\
\mathcal{L}_{lpips}& =\operatorname{LPIPS}\left(I_{pred}, I_{gt}\right).
\end{aligned}
\end{equation}


Moreover, to better preserve fine-grained mouth details (e.g., teeth), we additionally apply LPIPS loss focused on the mouth region, guided by a face parsing mask $M_{\mathrm{mouth}}$:
\begin{equation}\mathcal{L}_{m-lpips} = \operatorname{LPIPS}(I_{pred} \odot M_{mouth}, I_{gt} \odot M_{mouth}).\end{equation}

Finally, to stabilize Gaussian attributes, we introduce L2 regularization on position and scale:
\begin{equation}
\mathcal{L}_{xyz} = \| P_{\mathrm{pred}} - P_{\mathrm{init}} \|_2^2, \quad
\mathcal{L}_{scale} = \| S_{\mathrm{pred}} - S_{\mathrm{init}} \|_2^2.
\end{equation}

The overall training objective is a weighted combination of these terms:
\begin{equation}
\begin{aligned}
\mathcal{L} &= \lambda_{l 1} \mathcal{L}_{l 1}+\lambda_{ssim} \mathcal{L}_{ssim }+\lambda_{lpips } \mathcal{L}_{lpips} \\
  &+ \lambda_{mouth} \mathcal{L}_{m-lpips} + \lambda_{xyz} \mathcal{L}_{xyz} + \lambda_{scale} \mathcal{L}_{scale}. \\
\end{aligned}
\label{eqn:loss}
\end{equation}


\vspace{1mm}
\subsection{Efficient Refinement Strategy}

While the flexible reconstruction model already supplies an high-quality animatable avatar with detailed deformations, we introduce an efficient refinement strategy to achieve better consistency with the input images, especially for challenging regions like hair and clothing that exhibit substantial personal variations.

To enhance these personalized characteristics without compromising the model's inherent ability to generate dynamic effects, we optimize the parameters of our reconstruction model while keeping the dynamic UNet frozen. Additionally, gradients related to the mouth region are detached during fine-tuning to mitigate interference from complex intraoral shadows. Our fine-tuning process is supervised by both photometric and perceptual losses as
\begin{equation}
{\theta_E^{\star}} \longleftarrow \text{argmin}_{\theta_E} \mathcal{L}_{l1,ssim,lpips}\big(\mathcal{R}(\text{LBS}(G_{\theta_E}), \Theta), I_{gt}\big),
\end{equation}
where $\mathcal{R}(\cdot)$ denotes the 3DGS renderer and $G_{\theta_E}$ the reconstruction model.
Since our feedforward results already exhibit a considerable level of realism,  20 iterations (about 10 seconds) of finetuning are sufficient to improve personalization without compromising real-time rendering efficiency.

\section{Experiment}

\begin{figure*}[h]
    \centering
    \includegraphics[width=\linewidth]{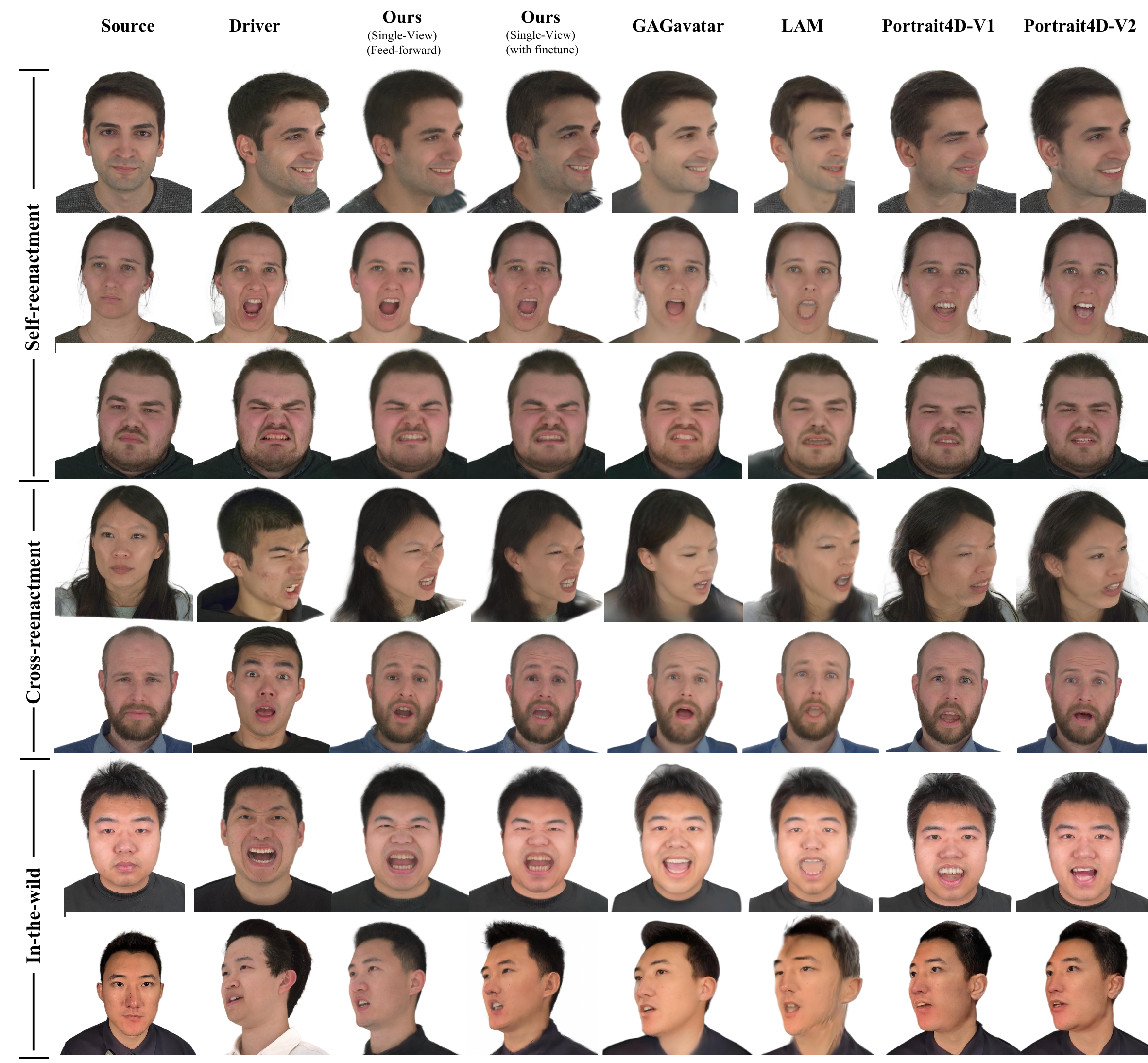}
    \vspace{-3mm}
    \caption{\textbf{Qualitative comparisons with baseline methods.} Our single-image feed-forward and finetuned results both outperform other methods in terms of 3D consistency and animation quality, especially on details like wrinkles or teeth. Please zoom in to see the details.}
    \vspace{-3mm}
    \label{fig:Comparison}
\end{figure*}

\textbf{Dataset.} We train our generalizable model using two comprehensive datasets: the NeRSemble~\cite{kirschstein2023nersemble} dataset and our self-captured FaceCap dataset. For training, we utilized 150 subjects from NeRSemble and 2000 subjects from FaceCap, while 16 subjects from NeRSemble and 16 subjects from FaceCap are used for comparison and evaluation. Due to our data distribution adjustment, we employ 26 expressions and 16 viewpoints per subject for training on the NeRSemble dataset, while for our FaceCap, we utilize 26 expressions and 30 viewpoints from 360 degree per subject for training, thus enable full head generation.

\vspace{1mm}
\noindent{\textbf{Training.} }
The whole hybrid model is trained on 16 80G GPUs end-to-end for about 4 days. Training uses the Adam optimizer with a learning rate of $3 \times 10^{-5}$, and loss weights are set as $\lambda_{l1}=1, \lambda_{ssim}=0.1, \lambda_{lpips}=0.2, \lambda_{mouth}=10, \lambda_{xyz}=0.01, \lambda_{scale}=1$. 

\vspace{1mm}
\noindent{\textbf{Inference Time.}}
For avatar creation, the forward encoding process takes approximately \textbf{0.4s} for four input images. For each driving expression, the process time of UNet forward pass, LBS and splatting is around \textbf{22 ms} on a single gpu, approximately \textbf{45 fps}. When refinement is applied, the process requires a mere \textbf{10s}, enabling immediate real-time animation of the avatar through our UNet.

\newcommand{\mrka}[1]{{\colorbox{red!30}{#1}}}
\newcommand{\mrkb}[1]{{\colorbox{red!20}{#1}}}
\newcommand{\mrkc}[1]{{\colorbox{red!10}{#1}}}
\begin{table*}
\begin{tabular}{c|cccccc|ccc}
\hline \multirow{2}{*}{Method} & \multicolumn{6}{|c|}{Self Reenactment} & \multicolumn{3}{c}{Cross Reenactment} \\
\cline{2-10}  & PSNR $\uparrow$ & SSIM $\uparrow$ & LPIPS $\downarrow$ & CSIM $\uparrow$ & AKD $\downarrow$ &  AED $\downarrow$ & CSIM $\uparrow$ & AKD $\downarrow$ & AED $\downarrow$ \\
\hline 
 LAM  & 17.8277 & 0.8031 & 0.2730 & 0.8213 & 5.7647 & 2.8077  & 0.8278 & 8.3807 & 4.8838 \\
 Portrait4D-v1  & 19.3746 & 0.8121 & 0.2410  & 0.8390 & 4.6719 & 2.3877 & 0.8399 & 8.0663 & 4.1993 \\
 Portrait4D-v2  & \mrkc{19.6279} & 0.8184 & \mrkc{0.2360}  & 0.8385 & 4.6910 & 2.4325 & 0.8389 & \mrkc{8.0622} & 4.2359 \\
 GAGAvatar  & 19.1667 & \mrkc{0.8283} & 0.2567 & \mrkc{0.8474}  & \mrkc{3.8730} & \mrkc{2.1175} & \mrkc{0.8479} & 8.2597 & \mrkc{4.0454} \\
\hline Ours(feed-forward) & \mrkb{21.1516} & \mrkb{0.8335} & \mrkb{0.2193}  & \mrkb{0.8490} & \mrkb{3.6518} & \mrkb{2.0502} & \mrkb{0.8501} & \mrkb{7.8726} & \mrkb{3.6415} \\
 Ours(with finetune) & \mrka{22.6313} & \mrka{0.8491} & \mrka{0.1833} & \mrka{0.8532}  & \mrka{3.4437} & \mrka{1.9127} & \mrka{0.8549} & \mrka{7.2426} & \mrka{3.5879} \\
\hline
\end{tabular}

\caption{\textbf{Quantitative comparisons on Nersemble test dataset.} In feedforward evaluations, our method surpasses other Gaussian-based single-image avatar generation approaches on every tested metric. Moreover, applying a subsequent finetuning stage yields additional gains in reconstruction fidelity and perceptual quality, further improving the realism and accuracy of generated outputs.}
\label{tab:nersemble_comparison} 
\end{table*}

\begin{figure*}[h]
    \centering
    \includegraphics[width=\linewidth]{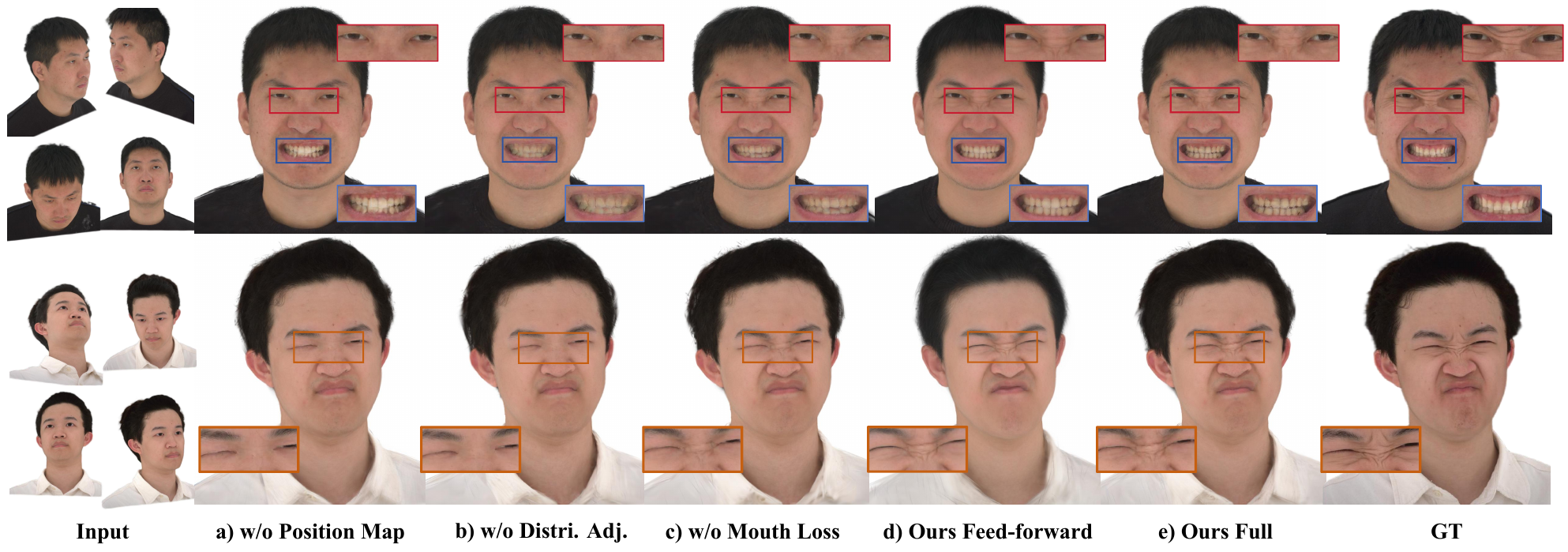}
    \vspace{-2mm}
    \caption{\textbf{Ablation study.} Using a position map as the driving signal (a) and performing Data Distribution Adjustment (b) substantially improve the control and realism of dynamic textures, such as oral cavities and wrinkles. Adding a mouth perception loss (c) increases the granular detail of teeth, yielding more complete dental appearance. Finally, applying a finetuning stage (d) further enhances consistency with the input image for widely varying person-specific attributes (e.g., hair, clothing).}
    \vspace{-2mm}
    \label{fig:Ablation}
\end{figure*}

\subsection{Comparison}

\noindent{\textbf{Baselines and Metrics.}}
We compare FlexAvatar with state-of-the-art one-shot avatar creation methods, including LAM~\cite{he2025lam}, GAGAvatar~\cite{chu2024gagavatar}, Portrait4D~\cite{deng2024portrait4d}, and Portrait4D-v2~\cite{deng2024portrait4dv2}, using their official implementations. Qualitative comparisons with the non-open-source Avat3R~\cite{Kirschstein2025avat3r} and HeadGAP~\cite{zheng2024headgap} are provided in the appendix Sec.~\ref{more-comparison}.

For self-reenactment (with ground truth), we evaluate image quality using PSNR, SSIM, and LPIPS, identity similarity via cosine distance of face-recognition features(CSIM) \cite{deng2019arcface}, and expression and pose accuracy via average expression distance (AED) and average keypoint distance (AKD) ~\cite{deng2019accurate}. For cross-identity reenactment (without ground truth), we only report CSIM, AED, and AKD.
To assess 3D consistency, we evaluate all models on four viewpoints from Nersemble’s test set. For fairness, all methods use a single input image for inference, fine-tuning, or FLAME estimation. As baselines differ in input and rendering pipelines, we align results using face-keypoint–based cropping before quantitative evaluation.

\begin{figure*}[t]
\vspace{-6mm}
    \centering
    \begin{subfigure}{0.33\textwidth}
        \centering       
        \includegraphics[width=\textwidth]{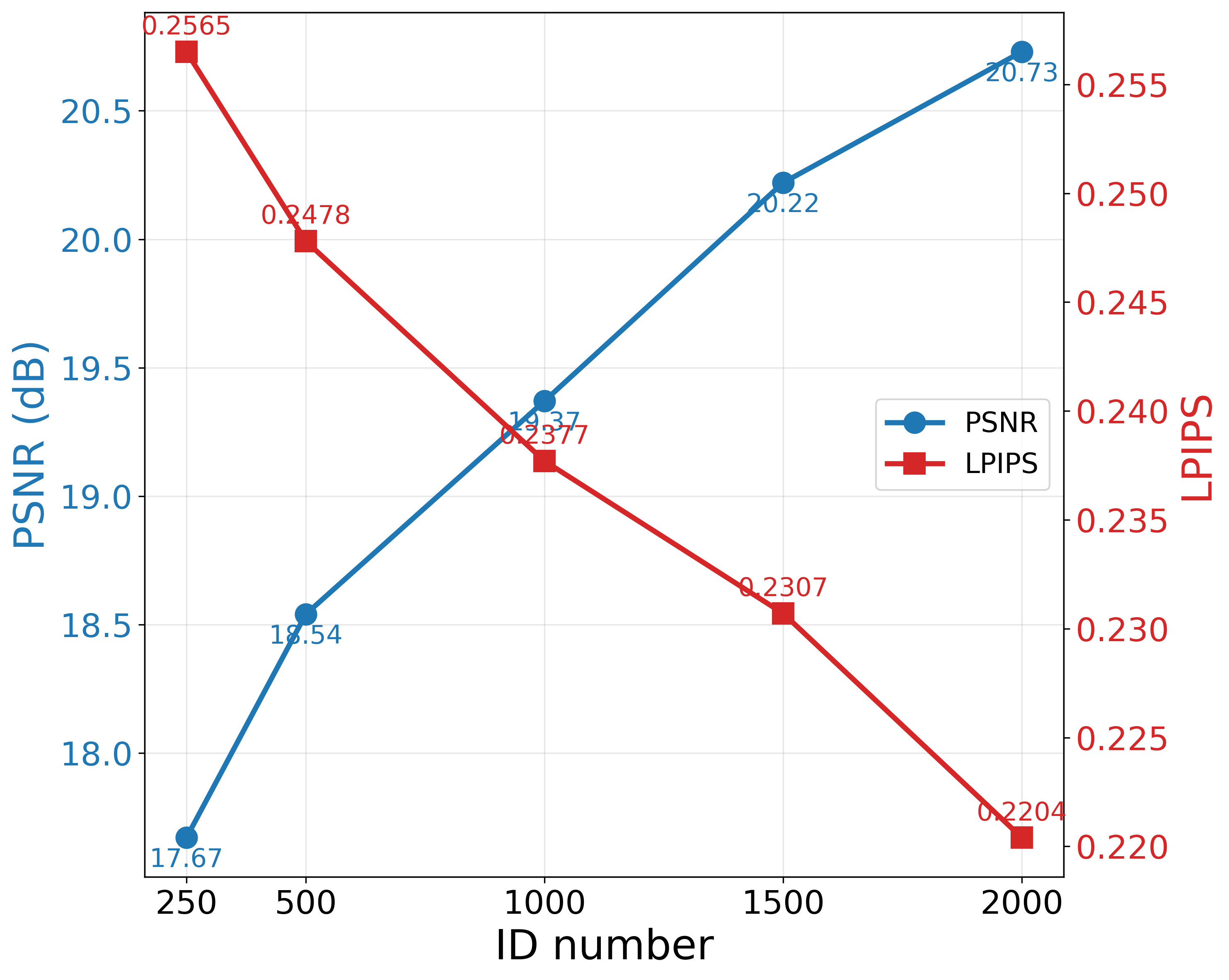}
        \caption{Feed-forward test results on FaceCap.}
        \label{fig:p}
    \end{subfigure}
    \begin{subfigure}{0.33\textwidth}
        \centering
        \includegraphics[width=\textwidth]{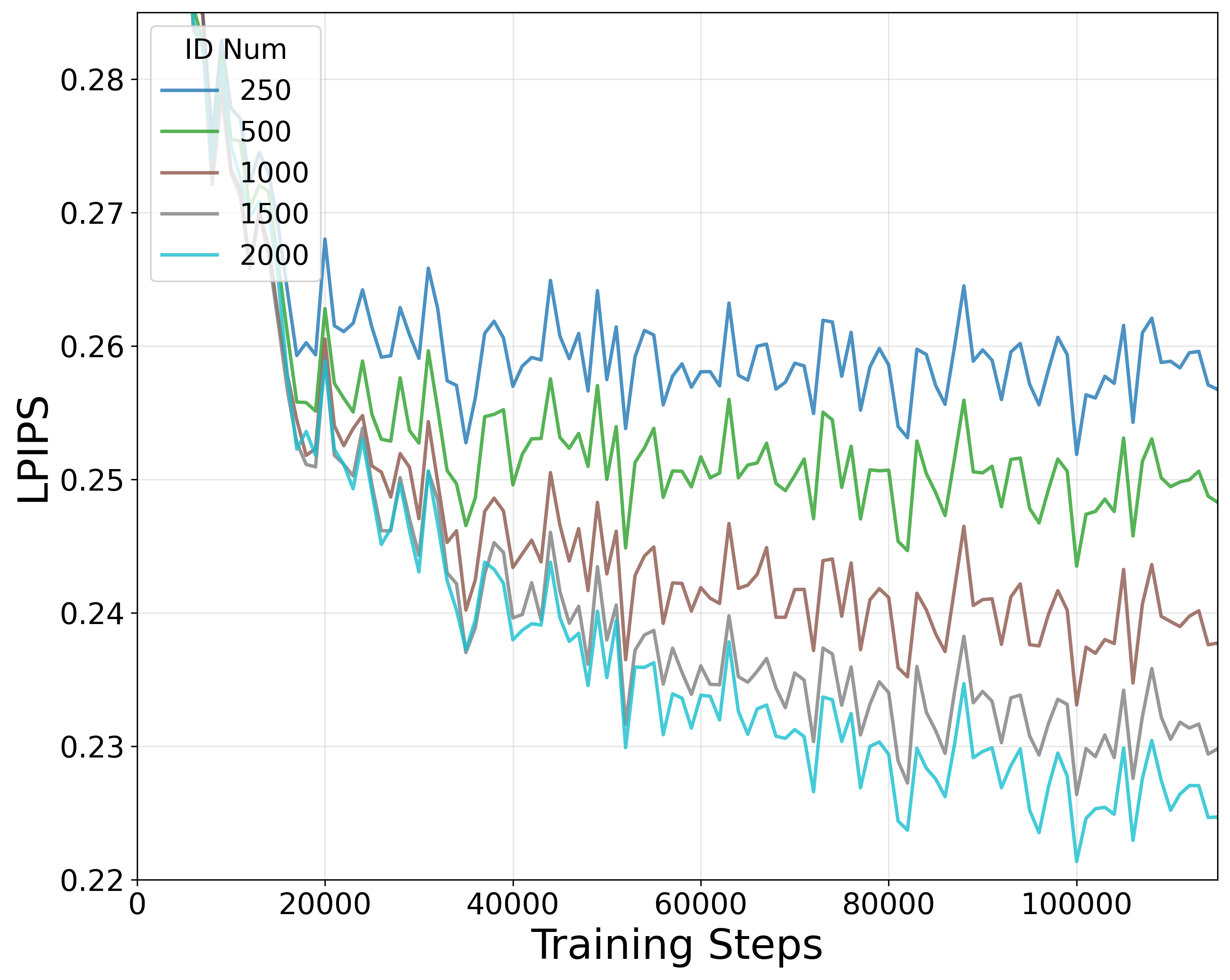}
        \caption{Model convergence speed.}
        \label{fig:p}
    \end{subfigure}
    \begin{subfigure}{0.33\textwidth}
        \centering
        \includegraphics[width=\textwidth]{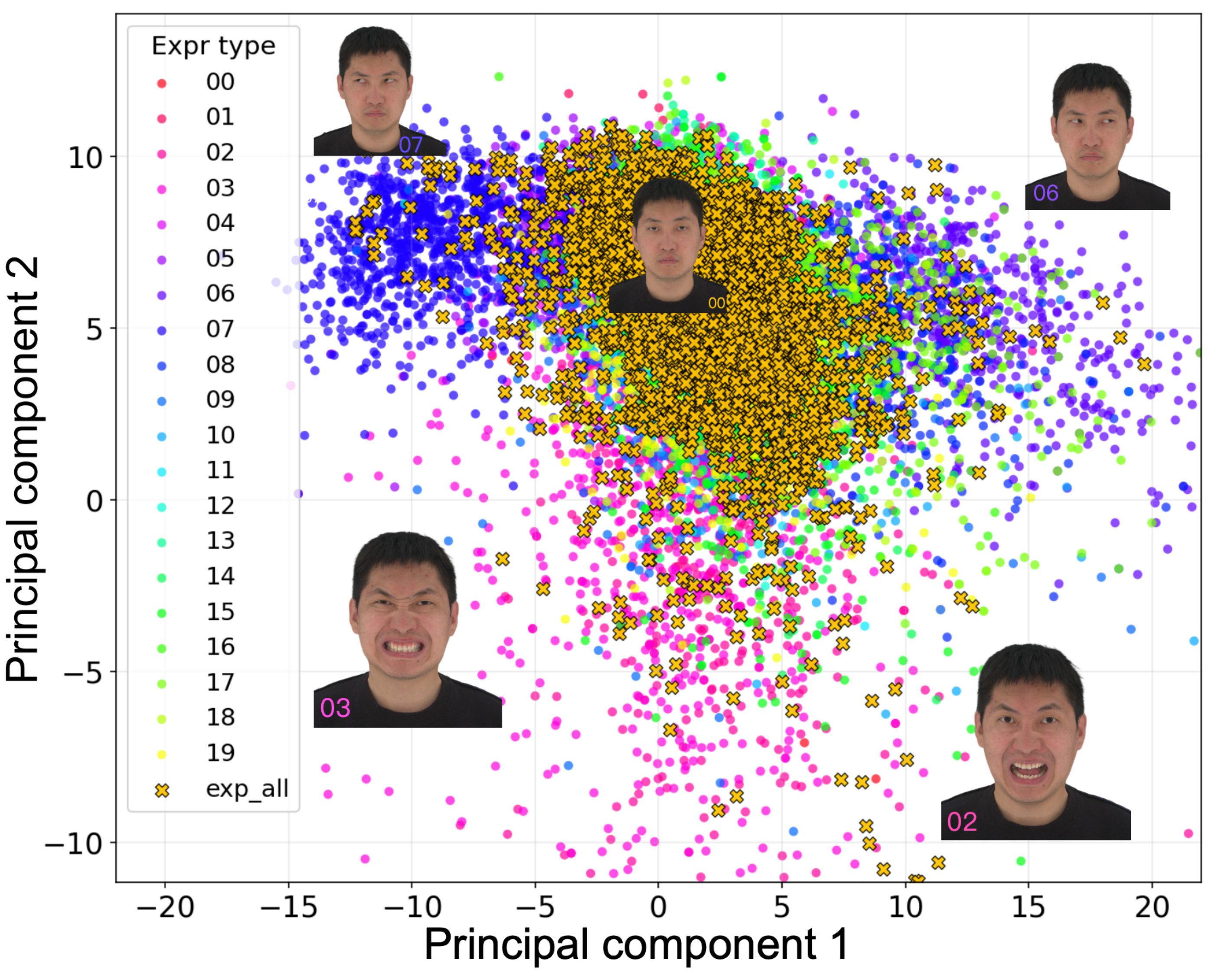}
        \caption{Expression PCA distribution.}
        \label{fig:pca_new}
    \end{subfigure}
     \vspace{-2mm}
    \caption{\textbf{Evaluation.} ~(a) The model's feed-forward test metrics show significant improvement as the training ID number increases. (b) Increasing the number of training identities led to a significant improvement in convergence speed. (c) A higher proportion of marginal expressions (eye rolling, mouth opening, frowning) is sampled after data distribution adjustment.}
    \label{fig:evaluation-fig}
    \vspace{-2mm}
\end{figure*}

\vspace{1mm}
\noindent{\textbf{Qualitative results.}}
Fig.~\ref{fig:Comparison} highlights two key advantages of our method. First, it achieves superior 3D consistency, while existing methods exhibit artifacts in side views (rows 4 and 7). Furthermore, our UNet-based animation produces realistic dynamics like wrinkles and teeth. Even without finetuning, our model preserves identity well in frontal views and surpasses baselines from the side, with finetuning further boosting identity consistency above all others.

\vspace{1mm}
\noindent{\textbf{Quantitative Comparison.}}
Tab. \ref{tab:nersemble_comparison} reports quantitative results on the Nersemble dataset for both self- and cross-reenactment. Across all metrics, our model consistently surpasses existing methods even before fine-tuning, and further achieves significant gains after fine-tuning.

\subsection{Evaluation}
\label{sec:Ablation}

\begin{figure}
        \centering
        \vspace{-2mm}
        \includegraphics[width=\linewidth]{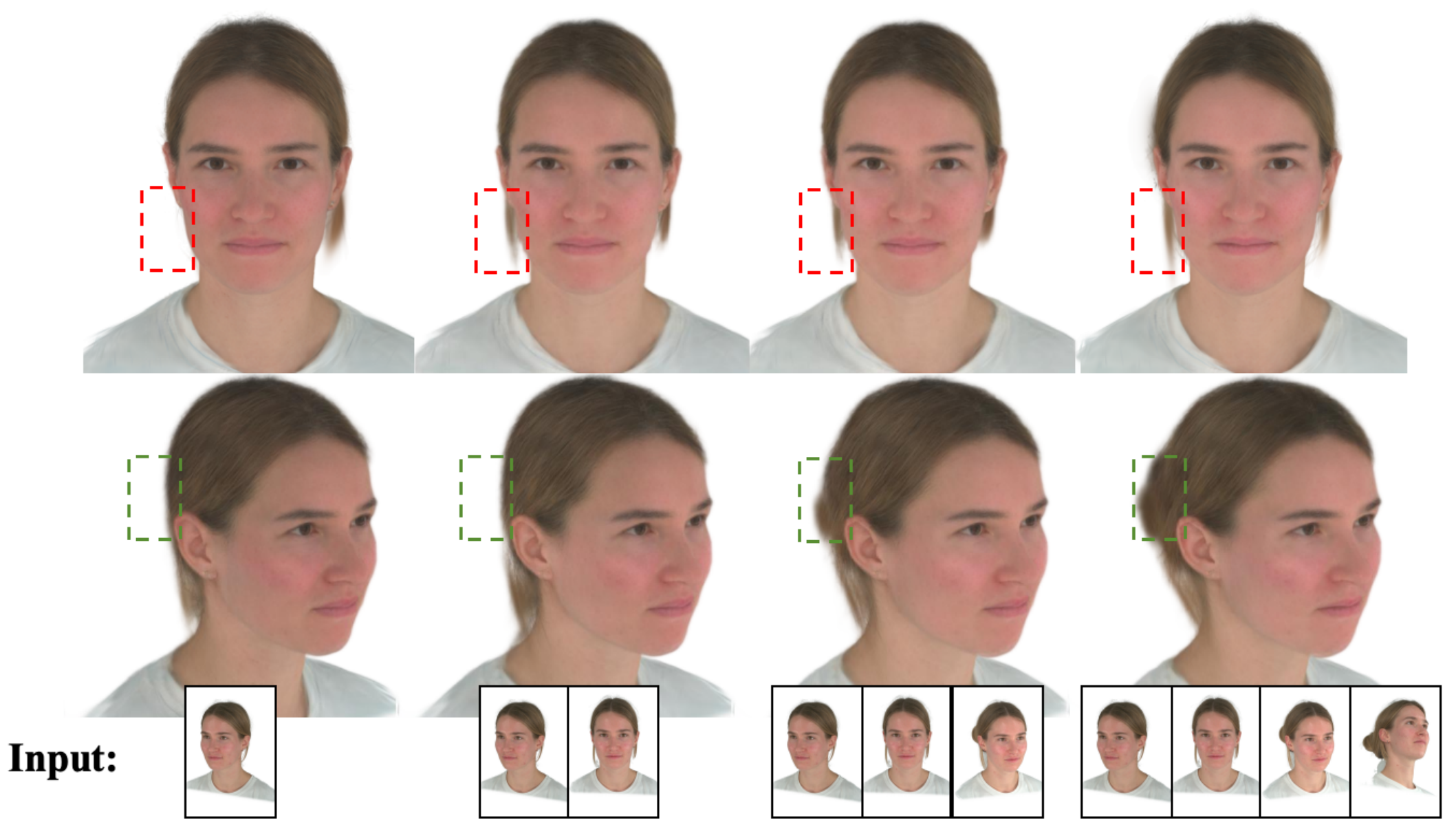}
        \caption{\textbf{Evaluation on the number of input images.} As the input images increase, the model's feed-forward results tends to learn more accurate shape and appearance information.} 
        \label{fig:num-input}
        \vspace{-2mm}
\end{figure}

\begin{figure}
        \centering
        \includegraphics[width=\linewidth]{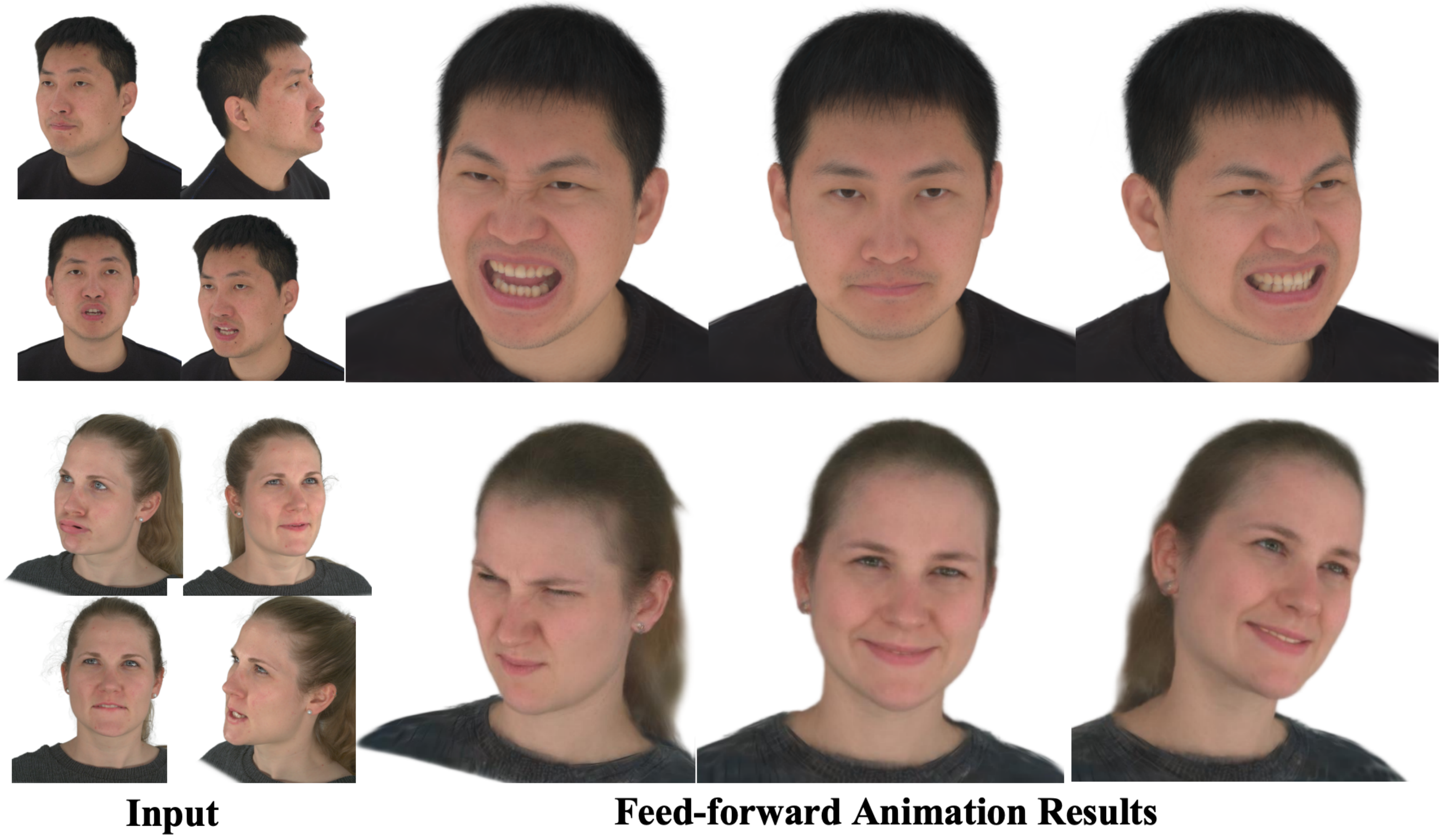}
        \vspace{-2mm}
        \caption{\textbf{Evaluation on the camera pose and expression of input images.} Our model can automatically adapt to different input viewpoints and expressions while maintaining good ID preservation capability and animation results.}
        \label{fig:exp-pose-free}
        \vspace{-2mm}
\end{figure}

\vspace{1mm}
\noindent{\textbf{{Ablation Study.}}
As shown in Fig.~\ref{fig:Ablation} and Tab.~\ref{tab:performance}, we ablate each component of our method. In (a), replacing the position map in the UNet input with a per-pixel concatenation of identity features and FLAME coefficients leads to a significant degradation in dynamic texture fidelity, producing smoother outputs with missing details like wrinkles and teeth. Removing the mouth-specific VGG perceptual loss in (c) demonstrates its crucial role in sharpening teeth and enhancing overall realism. The feed-forward results in (d) already show strong generation quality and high ID similarity even without finetuning, while (e) further improves consistency while preserving fine dynamic details.

\vspace{1mm}
\noindent{\textbf{Distribution Adjustment.}}
As mentioned in Sec. \ref{Decoding} and shown in Fig. \ref{fig:Ablation}(b), purely training with all data directly leads to degradation in dynamic wrinkles and teeth effects. Fig. \ref{fig:evaluation-fig}(c) compares the distributions of expression parameters before (the yellow cross label) and after (the colorful label 00-19) the process. We projected 10,000 randomly sampled parameters from both the full set and the anchor set into the 2D PCA space of the anchors. This comparison reveals that the anchor expressions provide a more uniform distribution, particularly for marginal expressions such as eye rolling, mouth opening, and frowning.

\vspace{1mm}
\noindent{\textbf{Training ID Number.}}
Fig.~\ref{fig:evaluation-fig}(a) and (b) shows how PSNR and LPIPS for our model’s feed‑forward results (w/o finetune) vary as the number of training identities increases, while Fig.~\ref{fig:id_num} demonstrates the qualitative changes. The experimental results indicate that the model’s performance improves markedly with more training identities, demonstrating a strong potential for scaling up.

\vspace{1mm}
\noindent{\textbf{Input Image Number, Expression and Camera Pose.}}
In Fig.~\ref{fig:num-input}, 
as more side‑view images are added, the feed‑forward output corrects estimates that are unobservable from a frontal view, showing that our network can accurately leverage information from different numbers of views. However, due to GPU memory constraints, we limit the model’s inputs to 4 images. Fig.~\ref{fig:exp-pose-free} demonstrates that our model robustly maintains accurate identity features and produces realistic driving effects despite variations in input viewpoints and expressions.

\begin{table}
        \centering
        \captionsetup{type=table}
        \begin{tabular}{lccc}
            \toprule
            Method & PSNR $\uparrow$ & LPIPS  $\downarrow$ & SSIM $\uparrow$ \\
            \midrule
            w/o Position Map  & 22.51 & 0.1845 & 0.8772 \\
            w/o Distri Adj & 22.74 & 0.1868 & 0.8745 \\
            w/o Mouth Loss & 23.10 & 0.1810 & 0.8890 \\
            Ours Full & \textbf{23.32} &\textbf{ 0.1797}& \textbf{0.8895} \\
            \hline
        \end{tabular}
        \vspace{-2mm}
        \caption{Quantitative Ablation Study on FaceCap.}
        \label{tab:performance}
        \vspace{-2mm}
\end{table}

\section{Limitation and Future Work}

While FlexAvatar achieves high-fidelity, real-time head avatars, it faces several limitations. Firstly, artifacts can occur with rare facial traits like glasses or hats due to their underrepresentation in training data. Secondly, non-head elements such as the body, clothing, and extreme complex hair are not modeled, limiting overall realism. In terms of lighting, the model shows moderate generalization; although the refinement helps, building fully relightable avatars remains a key future direction. Finally, scalability is both a challenge and an opportunity: avatar quality consistently improves with more 3D data, indicating that combining large-scale 3D datasets with abundant 2D data in a two-stage training scheme could significantly advance performance.
\section{Conclusion}

We present FlexAvatar, a framework for animatable 3D head avatar creation from single or sparse images, without camera poses or expression labels. It leverages a large reconstruction model to establish robust canonical representations. A lightweight UNet decoder, aided by a data distribution adjustment strategy, then captures fine-grained, expression-dependent dynamics. An optional test-time refinement module efficiently enhances fidelity for rare expressions and identity accuracy. Experiments show FlexAvatar surpasses state-of-the-art methods in 3D consistency, dynamic realism, and generalization.

\section*{Acknowledgments}
We sincerely thank Dongyang Zhao, Guidong Wang, Guanyi Chu, Kuan Tian, Xu Chang, Yang Zhao, Yifeng Wang, Zheng Shi and other volunteers for providing their portraits and assisting with data collection. Their support was essential to our research.

{
    \small
    \bibliographystyle{ieeenat_fullname}
    \bibliography{main}
}

\begin{appendices}
\clearpage
\setcounter{page}{1}
\maketitlesupplementary

\section{Implementation Details}
\label{sec:implementation}

\paragraph{Training settings.}
During training, we randomly configure 1 to 4 input images per batch to enhance model adaptability to varying expression inputs. All input images are sampled from random timesteps and random views, while four different viewpoints from distinct timesteps are randomly selected for supervision.

\paragraph{Data processing.}
For the Nersemble and FaceCap multi-view datasets, we employ the state-of-the-art multi-view FLAME \cite{li2017flame} estimation method VHAP \cite{qian2024vhap} for FLAME parameter estimation, use MODNet \cite{MODNet} for mask extraction, and utilize ParsingHuman \cite{Liu_2022_CVPR} for human parsing. For wild data, we adopt the advanced monocular FLAME estimation method Pixel3DMM \cite{giebenhain2025pixel3dmm} for FLAME tracking.

\paragraph{Network Architecture.}
Tab.~\ref{tab:hyperparameters} presents the architecture of our network and the specific settings of input/output parameters.

\begin{table}[htbp]
\centering
\begin{tabular}{lll}
\hline
& Hyperparameter & Value \\
\hline
\multirow{4}{*}{\begin{tabular}[c]{@{}l@{}} Image \\ Encoder \end{tabular}} 
&  Input image size & $512 \times 512$ \\
&  Input image number & $1$ to $4$ \\
&  Patch size & $16 \times 16$ \\
&  Output token dimension & $1024 \times 512$ \\
\hline
\multirow{2}{*}{\begin{tabular}[c]{@{}l@{}} Self-Attn \\ Block \end{tabular}} 
& Token dimension & $512$ \\
& Self-attn layers & 6 \\
\hline
\multirow{2}{*}{\begin{tabular}[c]{@{}l@{}}Cross-Attn \\ Block \end{tabular}} 
&  Head Query Token & $2500 \times 512$ \\
& Cross-attn layers & 6 \\
\hline
\multirow{4}{*}{\begin{tabular}[c]{@{}l@{}} Decoder \\ Block \end{tabular}} 
&  Upsample ratio & $\times 8$ \\
& Input dimension & $50 \times 50 \times 512$ \\
& ID feature map & $400 \times 400 \times 32$ \\
& Gaussian feature map & $400 \times 400 \times 14$ \\
\hline
\multirow{4}{*}{\begin{tabular}[c]{@{}l@{}}UNet \\ Block \end{tabular}} 
& Position map & $400 \times 400 \times 3$ \\
& Gaussian feature map  & $400 \times 400 \times 14$ \\
& Downsample layers  & 4 \\
& Upsample layers  & 4 \\
\hline
\end{tabular}
\caption{\textbf{Hyperparameters of our network architecture.}}
\label{tab:hyperparameters}
\end{table}

\paragraph{UV Map Structure.}
Fig.~\ref{fig:uv} illustrates the structure of the UV map we employed. On the left side of Fig.~\ref{fig:uv} is the layout of the FLAME vertices and faces in the UV coordinate system. To enhance the stability of the Gaussian points inside the oral cavity, we added a teeth region (the two rectangular areas at the top of the figure) to the original FLAME UV layout. This modification increases the number of vertices from the original 5,023 to 5,143. The right side of Fig.~\ref{fig:uv} shows a visualization of the UV Gaussian color map. The number of Gaussian primitives is set to n = 141,445, initialized from the UV map at a resolution of $400 \times 400$. Leveraging the regional UV masks provided by FLAME (e.g., for the mouth, face, hair) allows us to selectively fix the Gaussian primitives of certain areas during refinement.


\begin{figure}
        \centering
        \includegraphics[width=\linewidth]{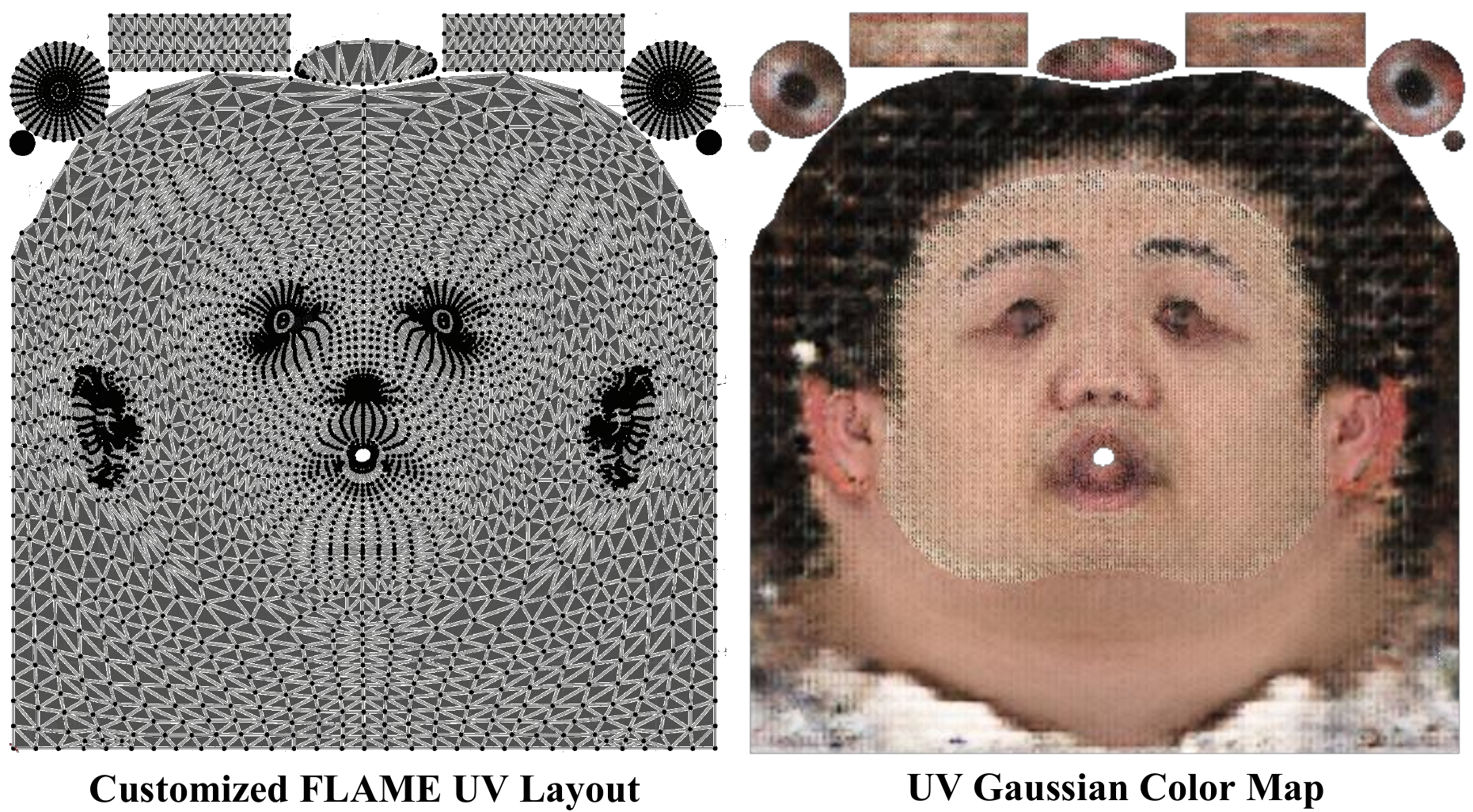}
        \caption{\textbf{UV Structure.} Our method use a $400 \times 400$ UV map structure to establish the mapping relationship between the Gaussian primitives and the FLAME model.}
        \label{fig:uv}
\end{figure}

\paragraph{Anchor Expressions.}
To identify the most expressive expressions, we employ a distribution-adjustment scheme that selects 20 anchor expressions for each person from the full dataset. A subset of the carefully curated anchor expressions is presented in Fig.~\ref{fig:Anchor}. These expressions include pronounced and representative actions (e.g., exaggerated mouth opening, deep furrowing of brows, and lateral mouth stretching). Incorporating these extremes provides the model with clear learning targets for handling challenging expressions, without compromising its performance on more common, subtle ones.

\section{Experiment Results}

\begin{figure*}[t]
    \centering
    \includegraphics[width=\linewidth]{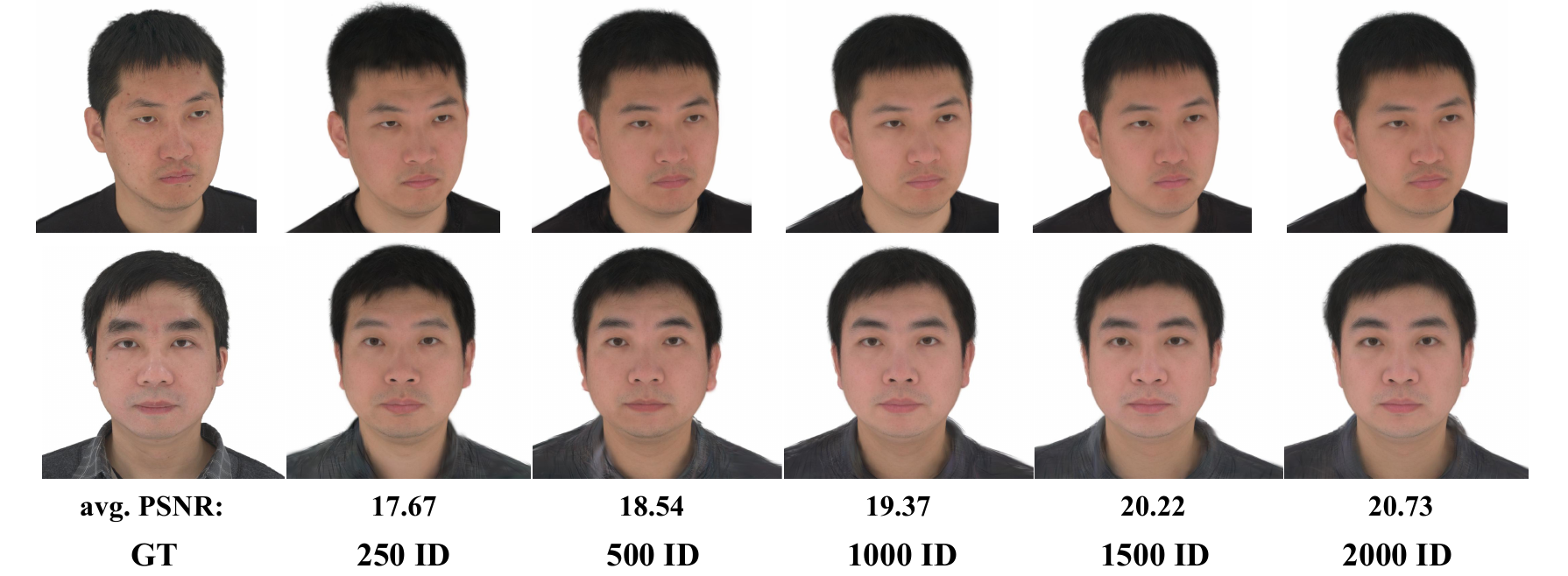}
    \vspace{-4mm}
    \caption{\textbf{Qualitative comparison on training id numbers.} The figure illustrates that our identity similarity increases as the number of training identities grows.}
    \vspace{-4mm}
    \label{fig:id_num}
\end{figure*}

\begin{figure*}[h]
    \centering
    \includegraphics[width=\linewidth]{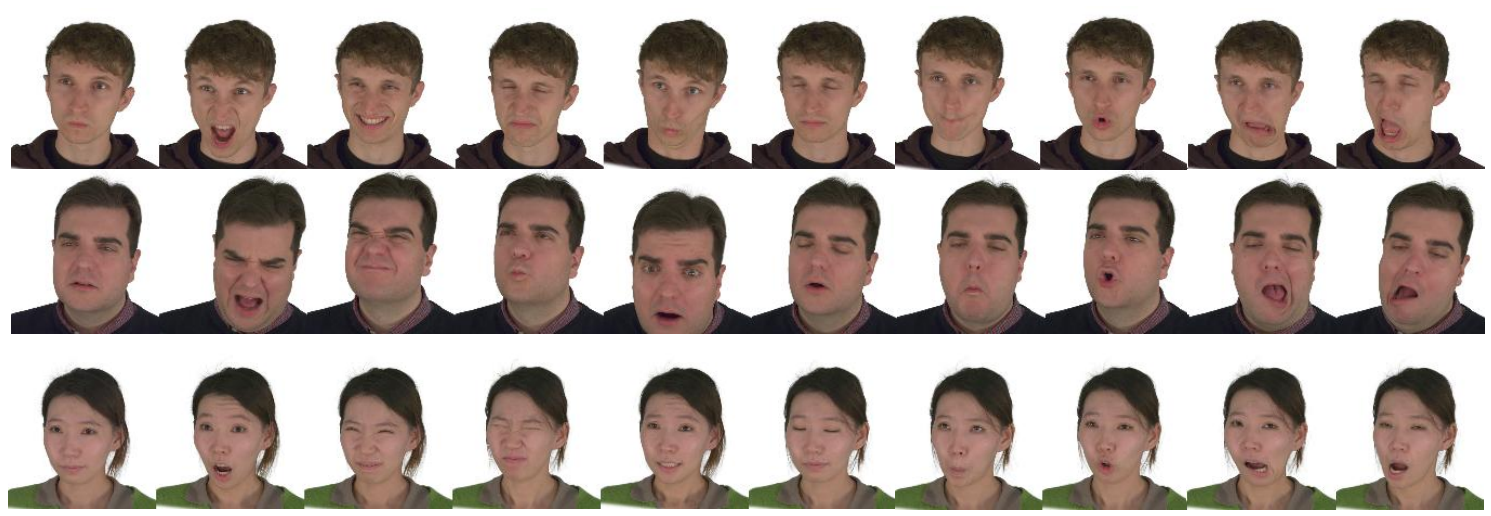}
    \vspace{-4mm}
    \caption{\textbf{Anchor Expressions.} Subset of our selected anchor expressions for training.}
    \vspace{-4mm}
    \label{fig:Anchor}
\end{figure*}

\paragraph{Effect of Number of Train Subjects.}
Fig.~\ref{fig:id_num} qualitatively demonstrates the improvement of our model with an increasing number of training identities. The results show a clear trend of rising identity consistency with the source image as the number of IDs grows. This trend validates the scalability of our approach. We anticipate that further increasing the scale of training data will continue to enhance model performance, ultimately allowing it to handle various challenging cases through a single feedforward pass.

\begin{figure*}[t]
    \centering
    \includegraphics[width=\linewidth]{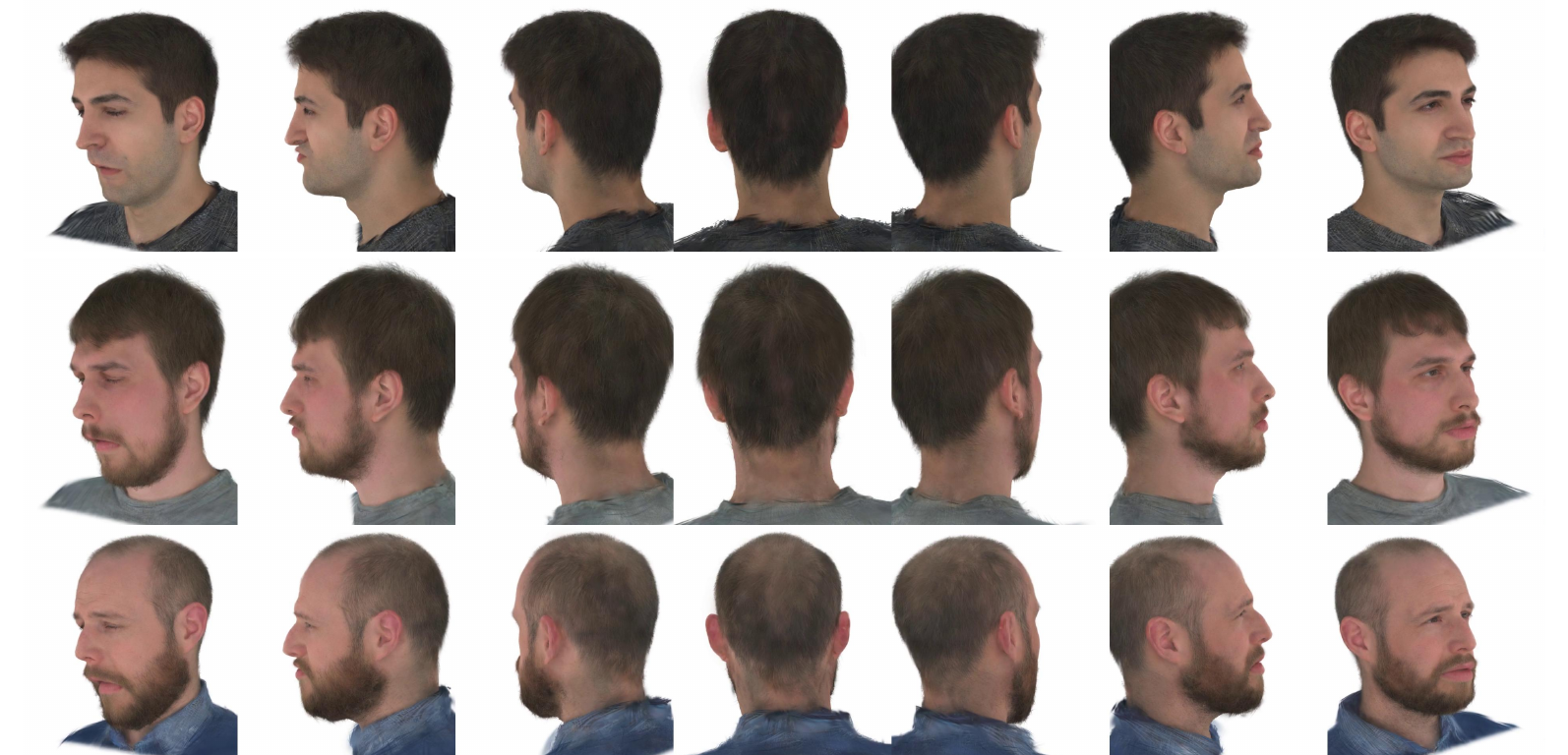}
    \vspace{-2mm}
    \caption{\textbf{Back head results.} 360-degree rendering of our head avatars.}
    \vspace{-6mm}
    \label{fig:back}
\end{figure*}

\begin{figure}[h]
    \centering
    \includegraphics[width=\linewidth]{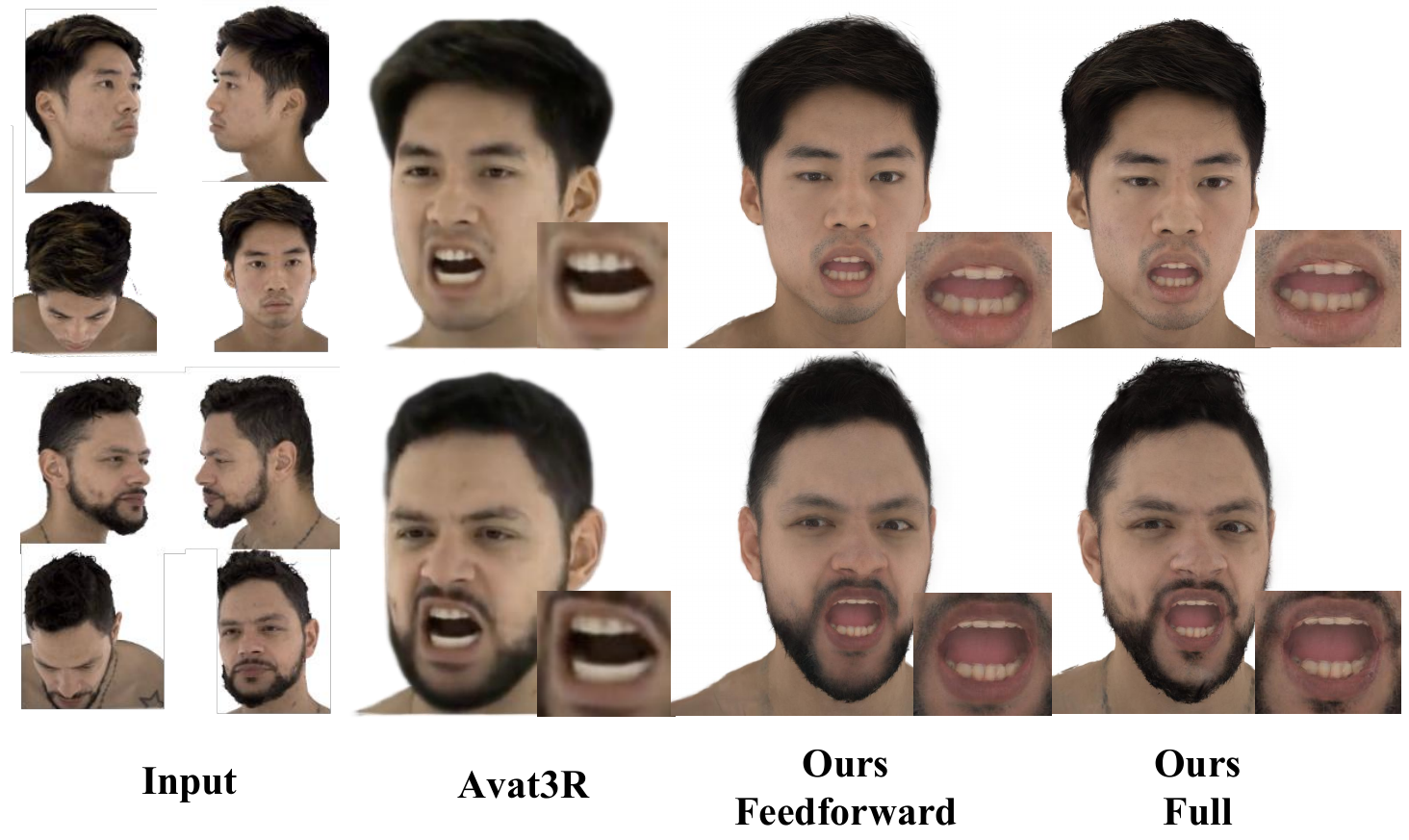}
    \vspace{-4mm}
    \caption{\textbf{Qualitative comparisons with Avat3r.} Our method outperforms Avatar3D in terms of clarity, similarity, and detail authenticity. Please zoom in to see details.}
    \vspace{-4mm}
    \label{fig:avat3r}
\end{figure}

\begin{figure}[h]
    \centering
    \includegraphics[width=\linewidth]{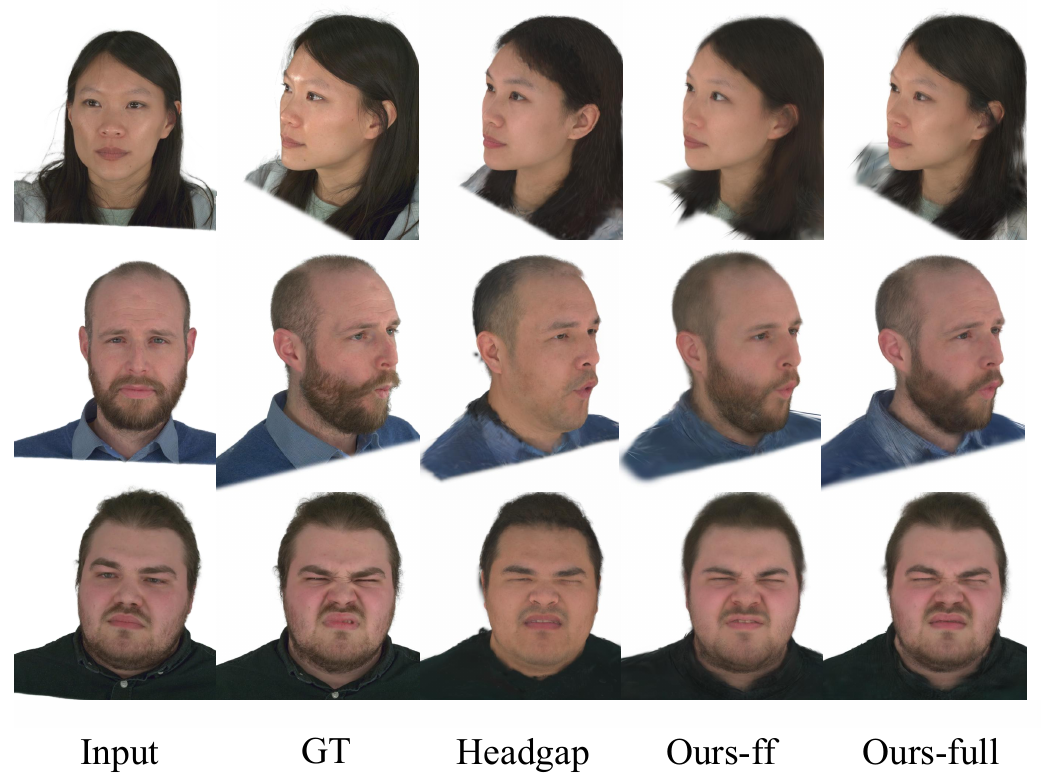}
    \caption{\textbf{Qualitative comparisons with Headgap~\cite{zheng2024headgap}.}}
    \label{fig:headgap}
    \vspace{-6mm}
\end{figure}

\paragraph{More Comparison.}
\label{more-comparison}
As shown in Fig.~\ref{fig:avat3r}, we further conducted a qualitative comparison between our method and Avat3R \cite{Kirschstein2025avat3r}, a state-of-the-art 4-view reconstruction method. Since Avat3R is not yet open-sourced, we evaluated our model using the same four input images showcased on their official website. As shown in our results, our method achieves superior performance in terms of identity similarity and finer details like teeth. It is worth noting that while Avat3R employs an expression latent for animation, granting it strong generative capabilities (e.g., for tongue generation), its cross-attention mechanism severely limits inference speed, reportedly reaching only 8 FPS. In contrast, our model, benefiting from a lightweight U-Net architecture, achieves a real-time driving speed of 45 FPS, which is crucial for interactive digital human applications.

\noindent Fig.~\ref{fig:headgap} presents qualitative comparisons of self-reenactment results between our method (under single-image input settings) and HeadGAP~\cite{zheng2024headgap}'s inversion + fine-tuning approach trained on our FaceCap dataset.
HeadGAP~\cite{zheng2024headgap} is a method that performs latent code inversion based on a prior model, followed by input-view refinement. The quantitative metrics for HeadGAP are as follows: PSNR: 20.77, SSIM: 0.8343, LPIPS: 0.2210, CSIM: 0.8489, AKD: 3.6739, AED: 2.1148. These results are comparable to our feedforward output but significantly inferior to our refined results. The comparison demonstrates that, although both methods employ input-view refinement, our approach achieves superior rationality and consistency in the refined outputs compared to HeadGAP.

\paragraph{More Results.}
Fig.~\ref{fig:back} showcases our full-head rendering results. Benefiting from the full-head data in the FaceCap dataset and our UV-aligned design, our model successfully generates complete head reconstructions. Fig.~\ref{fig:Self} presents additional self-reenactment results, while Fig.~\ref{fig:cross1} and Fig.~\ref{fig:cross2} display more cross-reenactment examples. These results collectively demonstrate our model's strong adaptability to a wide range of expressions. It performs well not only on in-domain datasets like FaceCap and Nersemble but also generalizes effectively to in-the-wild images captured by mobile phones. This provides a practical solution for creating high-quality avatars from a few uncalibrated images.

\paragraph{Failure Case.}
As discussed in our limitation section, the current system still exhibits several unresolved challenges.
Owing to the lack of relevant training data and the absence of specialized geometric design, our model struggles with examples featuring very fluffy long hair or eyeglasses. We posit that employing additional, dedicated Gaussian modeling for these specific areas could be a viable solution. We leave the investigation of these challenges for future work.

\section{Social impact}
Our work presents a paradigm shift for applications reliant on realistic digital humans. By streamlining avatar creation from minimal input, it democratizes high-quality character generation for VR, gaming and telehealth. This efficiency paves the way for scalable, practical, and engaging avatar applications across sectors.

\begin{figure*}[h]
    \centering
    \includegraphics[width=\linewidth]{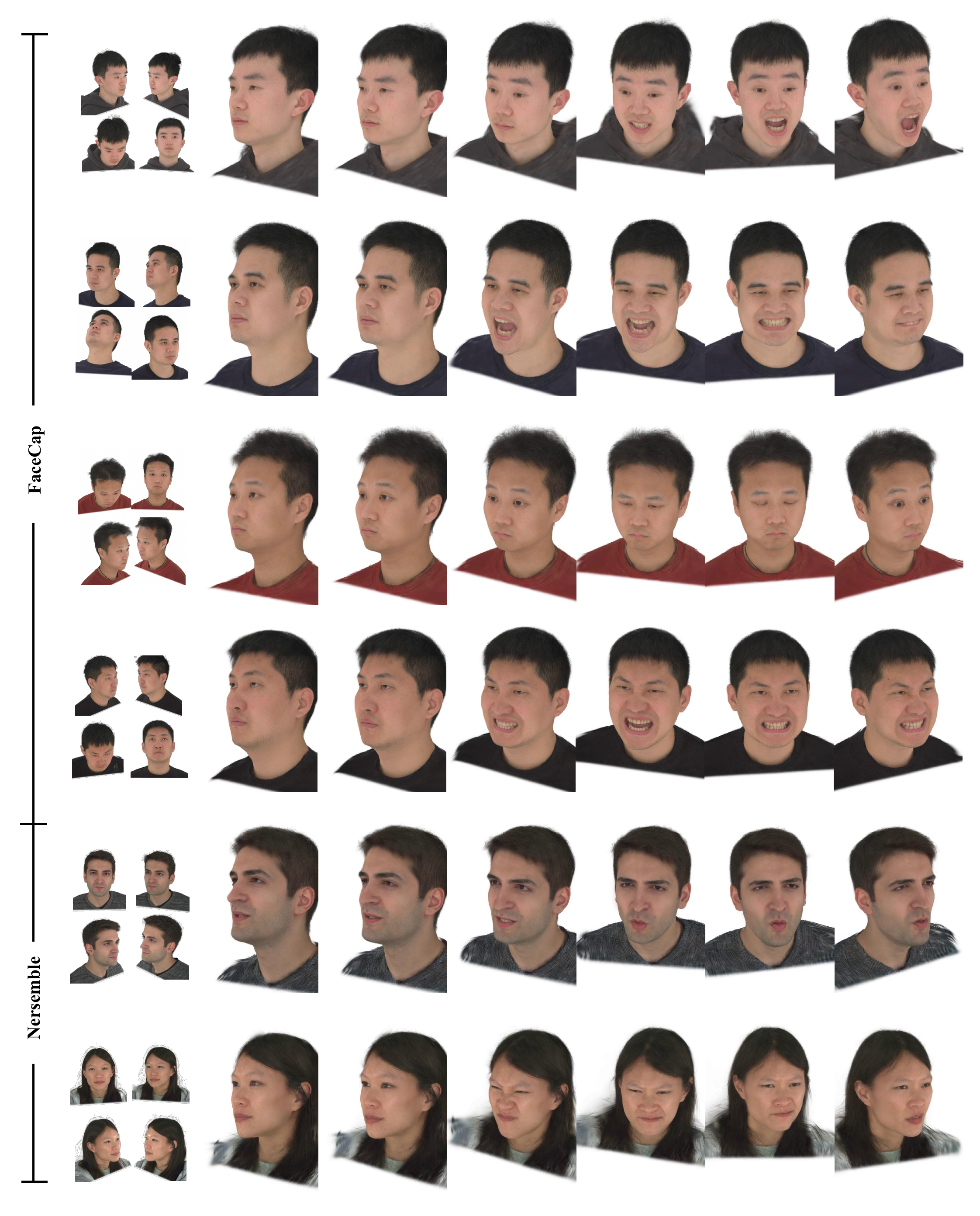}
    \vspace{-2mm}
    \caption{\textbf{Additional self-reenactment results.}}
    \vspace{-2mm}
    \label{fig:Self}
\end{figure*}

\begin{figure*}[h]
    \centering
    \includegraphics[width=\linewidth]{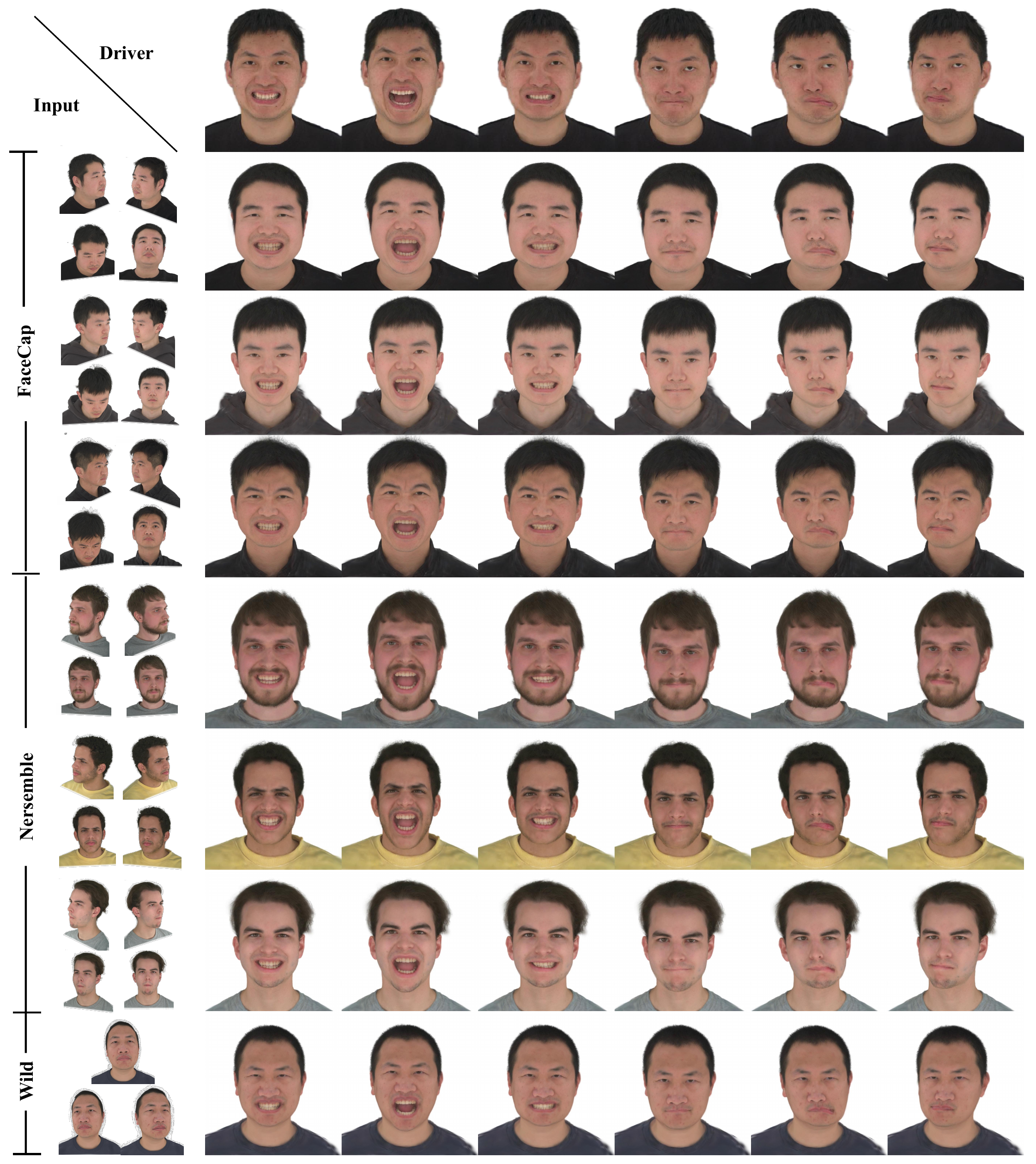}
    \vspace{-2mm}
    \caption{\textbf{Additional cross-reenactment results.}}
    \vspace{-2mm}
    \label{fig:cross1}
\end{figure*}

\begin{figure*}[h]
    \centering
    \includegraphics[width=\linewidth]{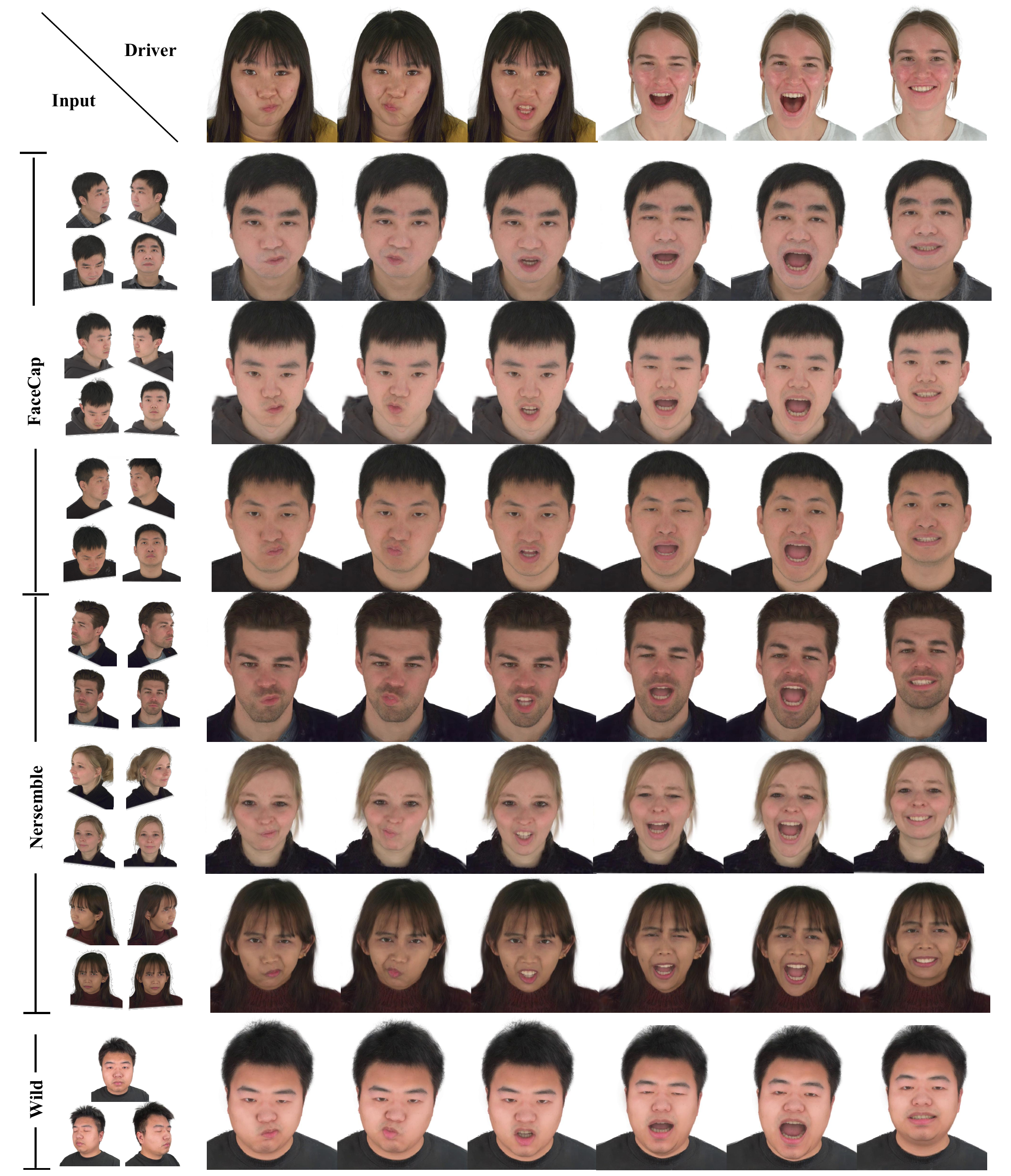}
    \vspace{-2mm}
    \caption{\textbf{Additional cross-reenactment results.}}
    \vspace{-2mm}
    \label{fig:cross2}
\end{figure*}

\end{appendices}


\end{document}